%% file: acl_latex.tex
\documentclass[11pt]{article}

\usepackage[final]{acl}

\usepackage{times}
\usepackage{latexsym}

\usepackage[T1]{fontenc}

\usepackage[utf8]{inputenc}

\usepackage{microtype}

\usepackage{inconsolata}

\usepackage{graphicx}

\input{math_commands.tex}

\usepackage{hyperref}
\usepackage{url}

\usepackage{subcaption}

\usepackage{booktabs}
\usepackage{multirow}
\usepackage{xcolor}
\usepackage{colortbl}
\usepackage{amssymb}
\usepackage{minted}

\usepackage{tikz}

\definecolor{keywords}{RGB}{255,0,90}
\definecolor{comments}{RGB}{0,0,113}
\definecolor{red}{RGB}{160,0,0}
\definecolor{green}{RGB}{0,150,0}

\definecolor{bananayellow}{rgb}{1.0, 0.88, 0.21}

\usepackage{wrapfig}
\usepackage{lipsum} 

\title{Exploring Solution Divergence and Its Effect on Large Language Model Problem Solving}

\author{
 \textbf{Hang Li},
 \textbf{Kaiqi Yang},
 \textbf{Yucheng Chu},
 \textbf{Hui Liu},
 \textbf{Jiliang Tang}\thanks{Corresponding author.}
\\
 Michigan State University,
\\
 \texttt{\{lihang4,kqyang,chuyuch2,liuhui7,tangjili\}@msu.edu} \\
}

\begin{document}
\maketitle
\begin{abstract}
Large language models (LLMs) have been widely used for problem-solving tasks. Most recent work improves their performance through supervised fine-tuning (SFT) with labeled data or reinforcement learning (RL) from task feedback. In this paper, we study a new perspective: the divergence in solutions generated by LLMs for a single problem. We show that higher solution divergence is positively related to better problem-solving abilities across various models. Based on this finding, we propose solution divergence as a novel metric that can support both SFT and RL strategies. We test this idea on three representative problem domains and find that using solution divergence consistently improves success rates. These results suggest that solution divergence is a simple but effective tool for advancing LLM training and evaluation. Code is available at \url{https://github.com/dse-ai-edu/solution-divergence.git}.
\end{abstract}

\section{Introduction}
\input{introduction}

\section{Related Work}
\input{related}

\section{Preliminary Study}
\label{sec:preliminary}
\input{preliminary}

\section{Divergence Fused Fine-tuning Methods}
\label{sec:fine-tune}
\input{method}

\section{Experiment}
\label{sec:experiment}
\input{experiment}

\section{Conclusion}
\input{conclusion}

\section*{Limitations}

In this work, we introduce a solution divergence metric and focus on investigating its effectiveness in characterizing LLM problem-solving behavior. Although our extensive experiments demonstrate the strong potential of this previously underexplored direction, several limitations remain. First, while edit distance and embedding cosine distance serve as simple and effective proxies for estimating solution divergence between paired solutions in our experiments, it is only an approximation. Language is inherently ambiguous, and more sophisticated divergence measures may better capture semantic and structural differences between solutions. Developing more precise and expressive divergence metrics is therefore an important direction for future research. Second, although we evaluate our approach on three representative problem-solving tasks to demonstrate the generality of the observed relationship, the correctness of solutions in these tasks is relatively easy to verify. In real-world scenarios, solution validity may be difficult or costly to assess, which could require additional processing or filtering of solution data before applying our method. Finally, our study primarily focuses on training-time analysis and experiments with small- to medium-sized models. Extending this framework to larger-scale models, as well as exploring techniques that directly leverage solution divergence during inference, remains an important and promising area for future work.

\bibliography{custom}

\newpage
\appendix
\input{appendix}

\end{document}

%% file: math_commands.tex
\usepackage{amsmath,amsfonts,bm}

\def\eqref#1{equation~\ref{#1}}

\def\1{\bm{1}}

\DeclareMathAlphabet{\mathsfit}{\encodingdefault}{\sfdefault}{m}{sl}
\SetMathAlphabet{\mathsfit}{bold}{\encodingdefault}{\sfdefault}{bx}{n}



%% file: introduction.tex
The rise of large language models (LLMs) and their remarkable general problem-solving capabilities have accelerated research on advanced artificial intelligence (AI) solutions across diverse domains, including science~\citep{ren2025towards}, finance~\citep{li2023large}, and education~\citep{wang2024large}. In particular, problems in STEM subjects such as mathematics~\citep{liu2024mathbench}, logic reasoning~\citep{parmar2024logicbench} and programming~\citep{coignion2024performance} have received significant attention, as their solutions can be objectively verified. A wide range of advanced algorithms have been proposed to improve LLMs’ problem-solving success, most of which focus on either expanding training datasets or applying supervised fine-tuning (SFT) with step-by-step annotated solutions~\citep{zhang2024scaling} or employing reinforcement learning (RL) with correctness-based rewards~\citep{ouyang2022training}.

While these approaches highlight the value of data in improving LLM performance, in this work we turn to an underexplored property shared across problem-solving datasets: the solution divergence, which refers to the presence of multiple viable solutions to a single problem. Studying solution divergence offers two key benefits in improving the models' performance. First, the majority of existing work in using SFT to boost LLM's performance on the task relies on generating or collecting new problems to augment training data, a process that is costly and labor-intensive due to the need for extensive cleaning and quality control~\citep{shen2024rethinking}. Although synthetic approaches such as question paraphrasing have been proposed~\citep{chen2024llm}, they often produce inconsistent quality and risk diverging from authentic problem distributions, limiting their effectiveness~\citep{chen2024unveiling}. By contrast, leveraging solution divergence allows us to enrich datasets using existing, authentic problems, thus avoiding these drawbacks.  Second, it is evident from cognitive science research that humans with larger repertoires of problem-solving strategies perform more effectively on complex tasks~\citep{siegler1998emerging}. Given the growing similarities between human and model problem-solving behaviors, such as step-by-step reasoning~\citep{wei2022chain}, we argue that studying the solution divergence potentially offers a new perspective for understanding LLM behavior from a cognitive-science point of view. Moreover, it provides new opportunities to improve LLM problem-solving performance. For instance, education research shows that fostering solution diversity in learners leads to better academic outcomes~\citep{caviola2018children}. By integrating solution divergence, we can analogously enhance LLMs.

Overall, our study is organized as follows. In \S~\ref{sec:preliminary}, we define the concept of solution divergence and evaluate it on three representative problem-solving datasets spanning mathematics, programming, and logical reasoning, examining its relationship with LLM performance across multiple models. In \S~\ref{sec:fine-tune}, we propose methods to incorporate solution divergence into both SFT- and RL-based fine-tuning paradigms. Finally, in \S~\ref{sec:experiment}, we present experiments on datasets from different domains and empirically demonstrate that integrating solution divergence enhances the problem-solving capabilities of LLMs.

%% file: related.tex
\paragraph{LLM Optimization.}

LLM optimization methods typically build on SFT and RL. SFT has proven effective across domains such as coding~\citep{roziere2023code}, mathematics~\cite{hendrycksmath2021,toshniwal2024openmathinstruct}, and general reasoning~\citep{yue2024mammoth2}, where curated datasets like Code Llama, OpenMathInstruct-2, and MAmmoTH2 yield substantial improvements. Reinforcement learning from human feedback (RLHF)~\citep{ouyang2022training} further established a widely used paradigm for aligning models to human preferences. More recently, group-based RL variants have been proposed to better capture sequence-level reasoning: GRPO~\citep{shao2024deepseekmath} introduces group-wise optimization to improve mathematical reasoning, DAPO~\citep{yu2025dapo} refines stability for long chain-of-thought training via dynamic sampling and token-level gradients, and GSPO~\citep{zheng2025group} adopts sequence-level clipping for greater efficiency. Our work complements these approaches by introducing solution divergence as an explicit signal, used both for selecting training samples in SFT and for designing diversity-aware reward functions in RL.

\paragraph{Data Diversity in LLM Training and Inference.}

Parallel lines of research emphasize the role of data and output diversity. At the training stage, diverse prompts and responses have been shown to enhance robustness and alignment~\citep{bukharin2023data,song2024scaling}, while synthetic data studies also report strong links between diversity and downstream generalization~\citep{chen2024diversity}. At inference, prompt ensembles, sampling strategies, and temperature scaling are commonly used to elicit multiple solution paths~\citep{kirk2023understanding}. Unlike these heuristic approaches, our framework formalizes diversity through a measurable solution divergence metric and integrates it directly into training objectives, unifying dataset-level diversity with inference-time diversity in a principled manner.

%% file: preliminary.tex
\subsection{Solution Divergence Definition}
\label{sec:div_define}

\begin{figure*}[!btph]
    \centering
    \includegraphics[width=.9\linewidth]{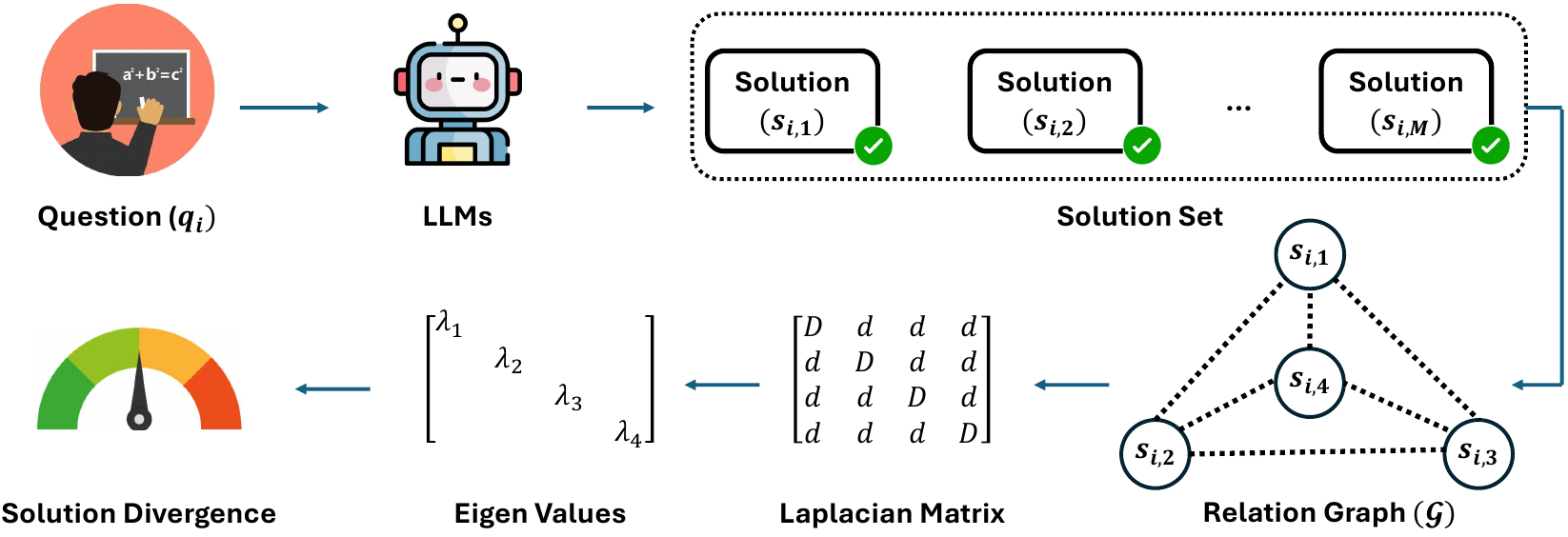}
    \caption{An Overview of the Solution Divergence Calculation}
    \label{fig:sd_diagram}
    \vspace{-3mm}
\end{figure*}

We consider a question dataset $\mathcal{Q}=\{q_n \mid n=1,\dots,N\}$ of size N. For each question $q_n$, the LLM ($\pi_{\theta}$) generates a solution set $\mathcal{S}_{q_n}=\{s_m \mid m=1,\dots,M\}$ of size M. The divergence between two solutions $s_i$ and $s_j$ is represented by $\delta_{i,j}$, capturing the degree of difference between them. For each question $q_n$, the overall divergence of its solution set $\mathcal{S}_{q_n}$ is denoted as $\zeta_{q_n}$. The model’s solution divergence over $\mathcal{Q}$ is written as $\zeta_{\pi}=\text{mean}(\zeta_{q_n})$. In cognitive science studies~\citep{caviola2018children}, the pairwise divergence $\delta_{i,j}$ is represented as a binary value $\{0,1\}$, where 0 indicates identical solutions and 1 indicates different solutions. These judgments are usually made by experts through manual review of solution pairs. Based on this, $\zeta_{q_n}$ is calculated by counting the number of unique solutions in the set of solutions. Formally, this can be expressed by constructing a weighted relation graph $\mathcal{G}$, where nodes correspond to solutions and edge weights correspond to their similarity, i.e., $1-\delta_{i,j}$ for $s_i$ and $s_j$. The number of connected components in $\mathcal{G}$ then yields $\zeta_{q_n}$. However, the scale of LLM-generated solutions in our study makes manual annotation infeasible. To address this limitation, we approximate $\delta_{i,j}$ using two types of proxy distance metrics: normalized string edit distance $d^{(e)}$ and sentence embedding cosine similarity $d^{(c)}$. More comparison details are provided in \S~\ref{sec:embedding}. Notably, due to substantial differences in solution formats across domains (e.g., mathematical derivations versus programming code), the effectiveness of each proxy metric varies by task. Therefore, for each problem, we empirically select the best-performing proxy metric. Exploring more advanced and unified divergence measures remains an important direction for future work.

To calculate $\zeta_{q_n}$, we derived it from the eigenvalues $\Lambda=\{\lambda_1,\dots,\lambda_M\}$ of the Laplacian matrix $L$ of the relation graph $\mathcal{G}$. This adjustment is necessary because the edge weights $\delta_{i,j}$ are non-binary. Inspired by spectral clustering~\citep{von2007tutorial}, where the magnitude of eigenvalues reflects the tightness of clusters in a relational graph, we propose two variants:

\vspace{-7mm}
\begin{equation}
\label{eq:div_def}
\zeta^{l}_{q_n} = M - \lambda_2,
\quad \zeta^{g}_{q_n} = M - \frac{1}{M}\sum_{i=1}^{M} \lambda_i .    
\end{equation}

\vspace{-2mm}

The local variant $\zeta^{l}_{q_n}$ is highly sensitive to weak local connections; even small changes in the smallest $\delta_{i,j}$ can substantially affect its value.  By contrast, the global variant $\zeta^{g}_{q_n}$ captures overall graph tightness from a global perspective. The solution divergence of an LLM is denoted as $\zeta^{l}_\pi$ and $\zeta^{g}_\pi$, respectively.  A schematic illustration of the solution divergence calculation is shown in Fig.~\ref{fig:sd_diagram}.

\subsection{Study Settings}
\label{sec:pre_setting}

\begin{table}[]
    \centering
    \resizebox{.49\textwidth}{!}{
        \begin{tabular}{|p{.1\textwidth}|p{.5\textwidth}|}
        \toprule
        Dataset & Question \\ \midrule
        Math-500 & A regular hexagon can be divided   into six equilateral triangles. If the perimeter of one of the triangles is   21 inches, what is the perimeter, in inches, of the regular hexagon? \\\midrule
        MBPP+ & Write a function to find   frequency of each element in a flattened list of lists, returned in a   dictionary. \\\midrule
        Maze & Given a 2D coordinate system   where both the x-axis and y-axis range from 0 to 10 (i.e., units 0, 1, ..,   10). Consider a point starting at position (0,0). The goal is to move this   point step by step to the target position: (8,4). During the moving, you cannot   pass the following position: (1,1), (3,4), (6,2). At each step, the point may   move only one unit right (r) or one unit up (u). Please provide one possible   sequence of moves to reach the destination. \\ \bottomrule
        \end{tabular}
    }
    \caption{Example problems from the datasets.}
    \label{tab:pre_example}
    \vspace{-5mm}
\end{table}

\paragraph{Datasets.}

To comprehensively study the relationship between solution divergence and LLM performance, we employ three representative problem-solving datasets: Math-500~\citep{lightman2023lets} (denoted as Math), MBPP+~\citep{liu2023your} (denoted as MBPP), and Maze. Math-500 and MBPP+ are well-established benchmarks for evaluating LLMs in mathematical problem solving and automatic code generation, respectively. In addition, we introduce Maze, a novel logical reasoning dataset developed for this study. Each  problem in Maze requires the model to identify a viable path from a given start point to an endpoint on a 2D coordinate grid while avoiding blocked areas. For every question across these datasets, correctness can be assessed objectively, and multiple valid solution paths may exist. To balance cost and efficiency, we sample 100 problems from the test split of each dataset for our experiments. Tab.~\ref{tab:pre_example} shows question examples from each dataset. Further details on all three datasets are provided in Appx.~\ref{appx:pre_data}.

\begin{figure*}[!btph]
     \centering
     \begin{subfigure}[b]{0.32\textwidth}
         \centering
         \includegraphics[width=\textwidth]{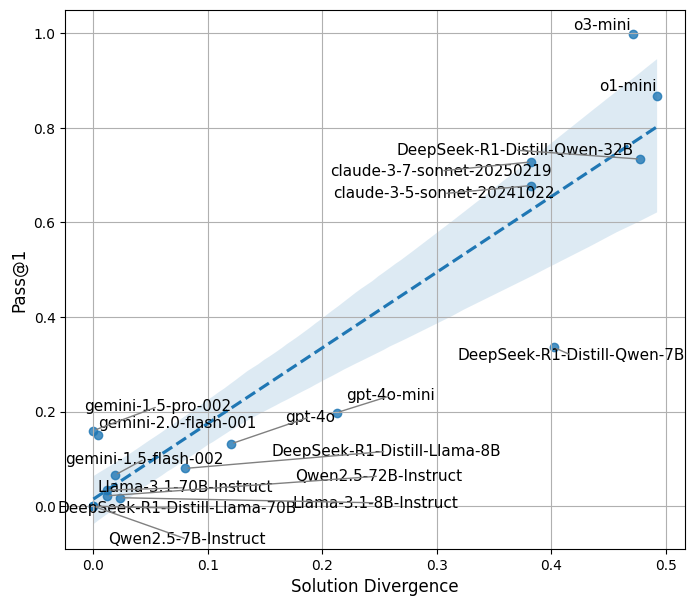}
         \caption{Pass@1 on Maze ($R^2$=.86)}
         \label{fig:pre_maze}
     \end{subfigure}
     \hfill
     \begin{subfigure}[b]{0.32\textwidth}
         \centering
         \includegraphics[width=\textwidth]{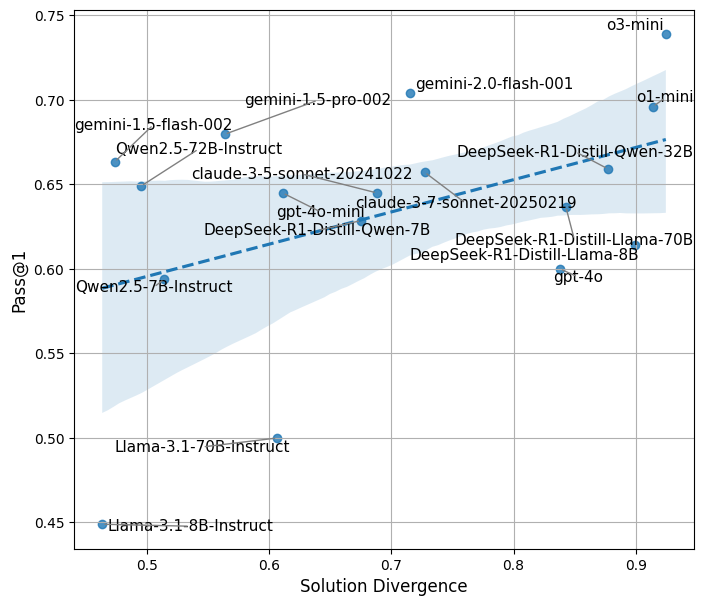}
         \caption{Pass@1 on Math ($R^2$=.19)}
         \label{fig:pre_math}
     \end{subfigure}
     \hfill
     \begin{subfigure}[b]{0.32\textwidth}
         \centering
         \includegraphics[width=\textwidth]{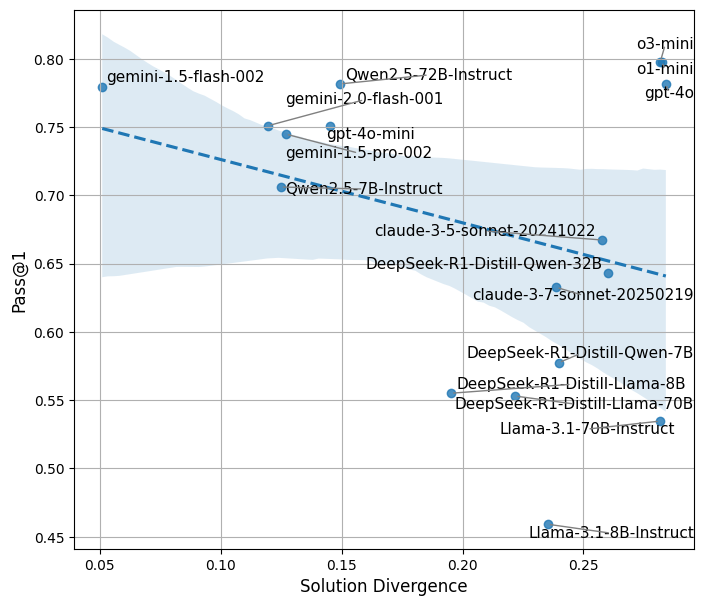}
         \caption{Pass@1 on MBPP ($R^2$=.09)}
         \label{fig:pre_mbpp}
     \end{subfigure}
     \begin{subfigure}[b]{0.32\textwidth}
         \centering
         \includegraphics[width=\textwidth]{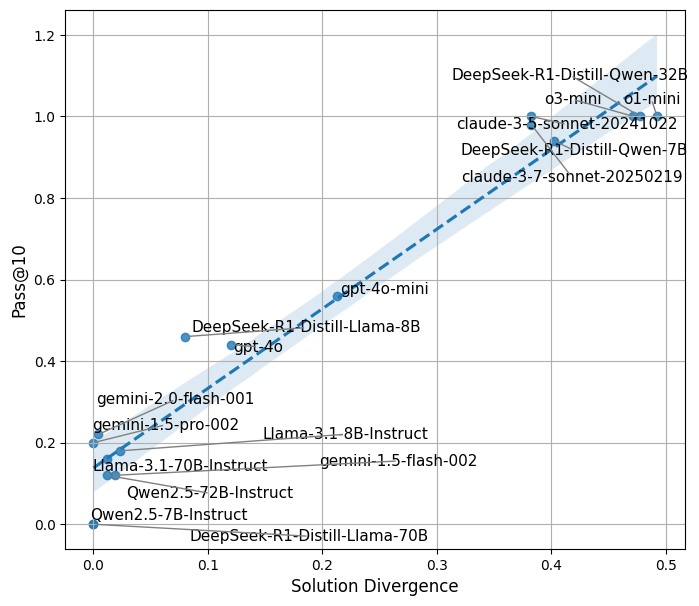}
         \caption{Pass@10 on Maze ($R^2$=.96)}
         \label{fig:pre_maze}
     \end{subfigure}
     \hfill
     \begin{subfigure}[b]{0.32\textwidth}
         \centering
         \includegraphics[width=\textwidth]{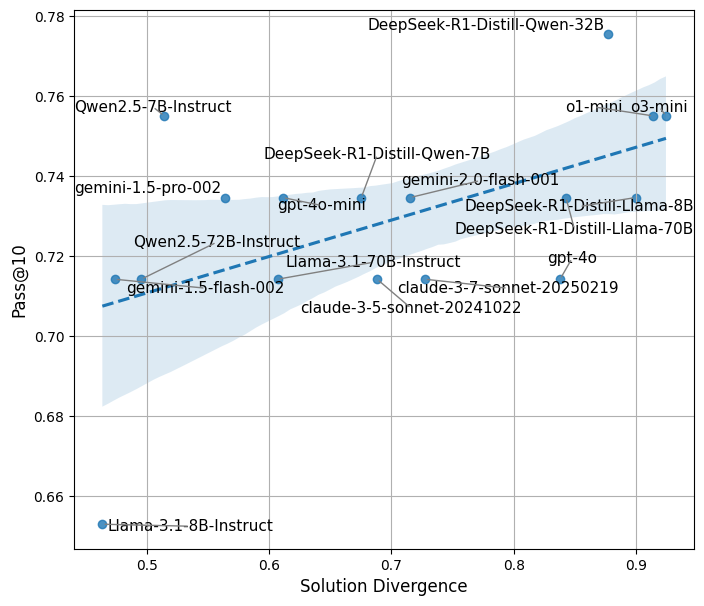}
         \caption{Pass@10 on Math ($R^2$=.31)}
         \label{fig:pre_math}
     \end{subfigure}
     \hfill
     \begin{subfigure}[b]{0.32\textwidth}
         \centering
         \includegraphics[width=\textwidth]{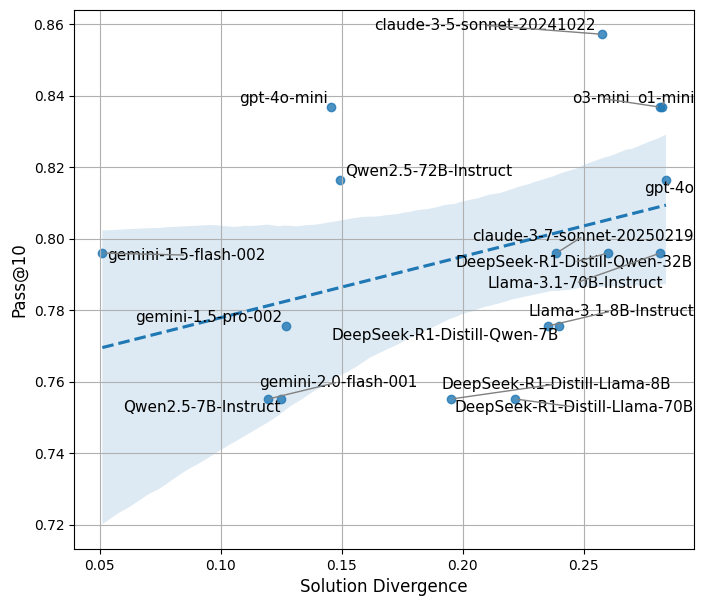}
         \caption{Pass@10 on MBPP ($R^2$=.14)}
         \label{fig:pre_mbpp}
     \end{subfigure}
     \caption{The Relationship of Solution Divergence ($\zeta^{g}_\pi$) to Success Rate (Pass@1 and Pass@10) in Maze, Math-500, and MBPP+ Datasets. For the Maze dataset, $\zeta^{g}_\pi$ is computed using the edit-distance-based metric $d^{(e)}$, while for Math-500 and MBPP+, it is computed using the embedding-based similarity metric $d^{(c)}$.}
     \label{fig:pre_passall}
     \vspace{-5mm}
\end{figure*}

\paragraph{Other Settings.}

We conducted experiments independently on the three datasets. In each experiment, different LLMs were treated as independent “testers,” analogous to participants in cognitive science studies, and were repeatedly prompted to solve the same problems, thereby generating the solution sets $\mathcal{S}_{q_n}$. To ensure independence between the evaluation of problem-solving performance and solution divergence, we randomly split the 100 sampled problems into two disjoint subsets. The first 50 problems were used to compute the divergence metric $\zeta_{\pi}$ for each model, while the remaining 50 problems were used to evaluate problem-solving performance, including the averaged single-attempt success rate (Pass@1) and the question-level ensembled success rate (Pass@10). Since divergence computation requires non-empty solution sets ($|\mathcal{S}{q_n}|>0$), we assigned $\zeta{q_n}=0$ for models that failed to generate any correct solutions within the allowed number of trials. To control for the influence of solution set size on divergence values, we required each model to produce the same number of correct solutions for every question. For models unable to generate a sufficient number of correct solutions, we applied random oversampling from the existing solution set to match the target size. Our study covers a broad spectrum of LLMs, including both open-source models (e.g., Meta Llama-3.1~\citep{touvron2023llama}, Alibaba Cloud Qwen-2.5~\citep{yang2024qwen2}) and closed-source models (e.g., OpenAI GPT-4o~\citep{bubeck2023sparks}, Anthropic Claude-3.5~\citep{TheC3}, and Google Gemini-1.5~\citep{team2023gemini}). To ensure fairness across models, we used the same inquiry prompt for all experiments and relied on each model’s default generation parameters. Additional details regarding the evaluated models and prompts are provided in Appx.~\ref{appx:pre_model} and Appx.~\ref{appx:pre_prompt}.

\subsection{Key Findings}
\label{sec:findings}

We present the relationship between $\zeta_\pi$ and the performance metrics Pass@1 and Pass@10 across the three datasets in Fig.~\ref{fig:pre_passall}. As shown in the plots, we observe a consistent positive correlation in five out of six cases between $\zeta_{\pi}^{g}$ and both performance metrics across all three tasks, supporting our hypothesis that LLM problem-solving performance is closely related to solution divergence. Moreover, the question-level performance metric Pass@10 exhibits a more stable positive correlation, as measured by the coefficient of determination ($R^2$), with $\zeta_{\pi}^{g}$ than Pass@1. This suggests that solution divergence may play a more significant role in improving question-level problem-solving performance. Finally, we analyze the relationship between solution divergence and Pass@1 under controlled question difficulty levels. Consistent with findings from cognitive science studies~\citep{caviola2018children}, we find that solution divergence is most informative for questions of moderate difficulty. More analyses and detailed results are provided in Appx.~\ref{appx:pre_line}.

%% file: method.tex
Based on the findings in \S~\ref{sec:findings}, we confirm a positive relationship between solution divergence ($\zeta^{g}_{\pi}$) and model problem-solving performance (Pass@1 and Pass@10) during inference. However, the effectiveness of solution divergence as a metric remains unverified in the training stage. Inspired by recognition science studies~\cite{siegler1998emerging}, which emphasize the benefits of fostering large repertoires of problem-solving strategies in children’s education, we propose two simple yet effective approaches that integrate solution divergence into existing training paradigms, i.e., SFT and RL, to further improve LLM performance. The following sections describe each method in detail.

\subsection{Dataset Divergence Metric}

The most straightforward way to leverage $\zeta_{\pi}$ for improving model performance during training is to use it as a criterion for data sample selection in the fine-tuning stage. The goal is to increase solution divergence by training the model on more diverse solutions. Data quality control is crucial in fine-tuning, as it directly affects both training efficiency and final performance~\citep{shen2024rethinking}. Building on this idea, our first proposed method for enhancing LLM problem-solving ability is to adopt $\zeta_{q_n}$ as a new metric for solution set selection. Specifically, for a given set of solutions $\mathcal{S}_{q_n}$ for the questions $q_n$, we compute its solution divergence to decide whether to add new solutions or remove low-value ones. The decision is guided by the change in $\zeta^{g}_{q_n}$ before and after modification: if the metric increases, the modification is accepted; otherwise, it is rejected. In this way, incorporating solution divergence into the supervised fine-tuning process ensures higher diversity within the training dataset, which in turn enhances the model’s problem-solving capability.

\subsection{Solution Divergence fused Reward}
\label{sec:rl_reward}

Another application of the solution divergence metric is its integration into reinforcement learning (RL) training for LLMs. Recent advances in reinforcement learning algorithms, such as GRPO~\citep{shao2024deepseekmath}, DAPO~\citep{yu2025dapo}, and GSPO~\citep{zheng2025group}, have demonstrated the effectiveness of using group-based success rewards, evaluated over a set of generated solutions $\mathcal{S}$\footnote{We omit the index in the following text for clarity} for each question, in removing the need for a separate value model, as required in the original RLHF framework~\citep{ouyang2022training}. However, these approaches focus solely on correctness-based rewards and neglect the solution divergence naturally present in group generation. To address this gap, we propose a novel divergence-augmented reward function $\mathcal{R}_\zeta(s_i,\mathcal{S})$, defined as:
\vspace{-1mm}
\begin{equation}
    \label{eq:reward_def}
    \mathcal{R}_\zeta = 
    \begin{cases}
    \left(\frac{|\mathcal{S}_c|}{|\mathcal{S}|}\right)^\alpha\cdot\frac{\sum_{s_j\in\mathcal{S}_c}\delta(s_i,s_j)}{|\mathcal{S}_c|}, & \text{if $v(s_i)=1$}, \\[12pt]
    -1, & \text{if $v(s_i)=0$}.
    \end{cases}
\end{equation}

\noindent where $s_i$ is the $i$-th generated solution, and $v(\cdot)$ is a verification function that returns 1 if $s_i$ is correct and 0 otherwise. $\mathcal{S}_c=\{s_i \mid v(s_i)=1\} \subseteq \mathcal{S}$ is the subset of correct solutions. $\delta(\cdot)$ denotes pairwise solution divergence calculation function, $|\mathcal{S}|$ is the number of sampled solutions, and $\alpha \in \mathbb{R}$ is a scaling hyperparameter. Compared with binary correctness-based rewards, $\mathcal{R}_\zeta$ incorporates pairwise solution divergence into reward computation for correct solutions. The leading term $(|\mathcal{S}_c|/|\mathcal{S}|)$ calculates the average success rate of the solution set and serves to balance correctness and diversity: when the success ratio is low, the reward emphasizes correctness; when success is high, it shifts attention toward divergence. The hyperparameter $\alpha$ controls the sensitivity of this balance. Intuitively, while classical binary success-based rewards treat all correct answers equally, $\mathcal{R}_\zeta$ reweights correct solutions according to their divergence. In standard RL training, optimization tends to concentrate on the dominant cluster of similar solutions, giving less influence to correct but more creative or diverse answers. By reweighting samples based on solution divergence, $\mathcal{R}_\zeta$ provides a more balanced learning signal between dominant and inventive solutions, enabling the model to learn from a broader range of behaviors. By summing over all solutions, we obtain the group reward for question $q_n$:
\vspace{-1mm}
\begin{equation}
\label{eq:reward}
    \mathcal{R}_{q_n} =\sum_{s_i\in\mathcal{S}}\mathcal{R}_\zeta(s_i,\mathcal{S}) \approx \left(\frac{|\mathcal{S}_c|}{|\mathcal{S}|}\right)^{\alpha-3}\cdot \zeta^g_{q_n} +|\mathcal{S}_c|-|\mathcal{S}|
\end{equation}

\begin{figure*}[!btph]
    \begin{equation}
    \label{eq:loss}
        \mathcal{J}(\theta)=\mathbb{E}_{(q,a)\sim\mathcal{D},\{s_i\}_{i=1}^{\mathcal{S}}\sim\pi_{\theta_{old}}(\cdot|q)}\left[\frac{1}{\sum_{i=1}^{|\mathcal{S}|}|s_i|} \sum_{i=1}^{|\mathcal{S}|}\sum_{t=1}^{|s_i|}\min\left(r_{i,t}(\theta)\hat{A}_{i,t},\text{clip}\left(r_{i,t}(\theta),1-\epsilon,1+\epsilon\right)\hat{A}_{i,t}\right)\right]
    \end{equation}
    \vspace{-10mm}
\end{figure*}


\noindent where $\zeta_{q_n}$ is the solution divergence for question $q_n$ we defined in \S~\ref{sec:div_define}. Full details of the simplification are provided in Appx.~\ref{appx:reward_simplify}. This formulation shows that the reward depends jointly on $\zeta_{q_n}$ and $|\mathcal{S}_c|$, encouraging the model not only to increase the number of correct solutions but also to diversify the solution set. For optimization, we adopt the Token-level Policy Gradient Loss proposed in DAPO, which alleviates the underweighting of long responses in the original GRPO loss. Furthermore, we remove the KL-divergence constraint to allow for broader exploration during RL training. The loss function is shown as Eq.~\ref{eq:loss}, where $r_{i,t}(\theta) = \frac{\pi_{\theta}(s_{i,t}|q,s_{i,<t})}{\pi_{\theta_{old}}(s_{i,t}|q,s_{i,<t})}$, and $\hat{A}_{i,t} = \frac{\mathcal{R}_i-\text{mean}(\{\mathcal{R}_i\}^{\mathcal{|S|}}_{i=1})}{\text{std}(\{\mathcal{R}_i\}^{\mathcal{|S|}}_{i=1})}$. Here, $s_{i,t}$ denotes the $t$-th token of solution $s_i$, $\theta$ is the parameter of the current policy model, and $\theta_{old}$ is the parameter of the reference model. The functions $\text{mean}$ and $\text{std}$ compute the mean and standard deviation of the rewards across the solution set $\mathcal{S}$. Finally, $\epsilon$ is the clipping hyperparameter that prevents excessively large updates and stabilizes RL training.

%% file: experiment.tex
In this section, we present the experimental details for the same three problem-solving tasks introduced in \S~\ref{sec:preliminary}. As in the previous section, we first describe the datasets and the preparation steps applied to each. We then provide details of the models and experimental settings. Next, we report results from both SFT- and RL-based algorithms, followed by ablation studies.

\subsection{Dataset}

Details of the problems posed to LLMs are provided in \S~\ref{sec:pre_setting}. Here, we focus on the preparation of the datasets used for subsequent supervised fine-tuning (SFT) and reinforcement learning (RL) training. For each task, we construct disjoint datasets for SFT ($\mathcal{D}_{\text{SFT}}$) and RL ($\mathcal{D}_{\text{RL}}$) through random sampling from the standard training split, ensuring that $\mathcal{D}_{\text{SFT}} \cap \mathcal{D}_{\text{RL}} = \varnothing$. Dataset sizes are determined based on sample availability, task difficulty, and computational constraints. Specifically, the RL training dataset is constructed by sampling a fixed number of questions together with their corresponding ground-truth answers from the original dataset. For SFT, since the experiments require not only correct answers but also multiple diverse valid solutions for each question, we employ advanced LLMs (e.g., GPT-4o, Gemini-2.5-Pro, and Claude-3.5-Sonnet) to generate at least 10 distinct correct solutions per question. For each question, we then select two subsets of solutions: the subset with the highest divergence is denoted as $\mathcal{D}_{\mathcal{S}}^{+}$, while the subset with the lowest divergence is denoted as $\mathcal{D}_{\mathcal{S}}^{-}$. By aggregating these subsets across all questions, we construct two versions of the SFT training dataset: a high-divergence dataset $\mathcal{D}_{\text{SFT}}^{+}$ and a low-divergence dataset $\mathcal{D}_{\text{SFT}}^{-}$. Additional details regarding dataset sizes, sampling procedures, and prompting strategies are provided in Appx.~\ref{appex:exp_data}.

\subsection{Settings}

We use four representative open-source LLMs in our experiments: Llama-3.2-1B, Llama-3.1-8B, Qwen2.5-1.5B, and Qwen2.5-7B. Following the commonly adopted performance improvement pipeline, each model is first trained with the task-specific SFT dataset, and then further refined with the RL dataset for additional performance gains. For the SFT stage, we train each model on both versions of the prepared datasets, $\mathcal{D}_{\text{SFT}}^{+}$ and $\mathcal{D}_{\text{SFT}}^{-}$. In the RL stage, we further train the models using the RL dataset $\mathcal{D}_{\text{RL}}$, starting from the two SFT-trained checkpoints. For both SFT and RL training, we tune the hyper-parameters including learning rate and the solution divergence balancing factor $\alpha$. The baselines are defined as follows: for SFT, models trained on $\mathcal{D}_{\text{SFT}}^{-}$ serve as the baseline for comparison with $\mathcal{D}_{\text{SFT}}^{+}$; for RL, we replace the divergence-fused reward function $\mathcal{R}_\zeta$ with the classical binary success-based reward function $\mathcal{R}_s$. For evaluation, we report Pass@1 (success rate for the top solution) and Pass@10 (success rate within the top 10 solutions. Task instruction prompts are identical to those used in our preliminary study; other details are provided in Appx.~\ref{appex:exp_setting}.

\begin{table*}[]
\caption{Problem-solving performance (Pass@1 and Pass@10, in \%) across three datasets for models tuned on $\mathcal{D}_{\text{SFT}}^-$ and $\mathcal{D}_{\text{SFT}}^+$. $\Delta = |\mathcal{D}_{\text{SFT}}^+ - \mathcal{D}_{\text{SFT}}^-|$, where {\color{green} green} indicates $\mathcal{D}_{\text{SFT}}^+$ exceeds that of $\mathcal{D}_{\text{SFT}}^-$, while {\color{red} red} indicates the opposite.}
\label{tab:sft}
\vspace{-2mm}
\centering

\resizebox{\textwidth}{!}{
\begin{tabular}{@{}cc|ccc|ccc|ccc|ccc@{}}
\toprule
\multicolumn{2}{c|}{Model} & \multicolumn{3}{c|}{Llama-3.2-1B} & \multicolumn{3}{c|}{Llama-3.1-8B} & \multicolumn{3}{c|}{Qwen2.5-1.5B} & \multicolumn{3}{c}{Qwen2.5-7B} \\ \midrule
\multicolumn{1}{c|}{Dataset} & Metric & $\mathcal{D}_{\text{SFT}}^-$ & \multicolumn{1}{c|}{$\mathcal{D}_{\text{SFT}}^+$} & $\Delta$ & $\mathcal{D}_{\text{SFT}}^-$ & \multicolumn{1}{c|}{$\mathcal{D}_{\text{SFT}}^+$} & $\Delta$ & $\mathcal{D}_{\text{SFT}}^-$ & \multicolumn{1}{c|}{$\mathcal{D}_{\text{SFT}}^+$} & $\Delta$ & $\mathcal{D}_{\text{SFT}}^-$ & \multicolumn{1}{c|}{$\mathcal{D}_{\text{SFT}}^+$} & $\Delta$\ \ \ \\ \midrule
\multicolumn{1}{c|}{\multirow{2}{*}{Maze}} & \ \ Pass@1 \ \ & \ \ 23.84 \ & \multicolumn{1}{c|}{\ 23.24\ \ } & \ \ \color{red}0.60 \ \ & \ \ 25.70 \ & \multicolumn{1}{c|}{\ 29.36\ \ } & \ \ \color{green}3.66 \ \ & \ \ 27.92 \ & \multicolumn{1}{c|}{\ 26.48\ \ } & \ \ \color{red}1.44 \ \ & \ \ 25.10 \ & \multicolumn{1}{c|}{\ 22.70\ \ } & \ \ \color{red}2.40\ \ \ \\
\multicolumn{1}{c|}{} & Pass@10 & 35.20 & \multicolumn{1}{c|}{43.80} & \color{green}8.60 & 36.80 & \multicolumn{1}{c|}{47.80} & \color{green}11.00 & 39.00 & \multicolumn{1}{c|}{43.60} & \color{green}4.60 & 36.00 & \multicolumn{1}{c|}{45.80} & \color{green}9.80\ \ \ \\ \midrule
\multicolumn{1}{c|}{\multirow{2}{*}{Math-500}} & Pass@1 & 22.88 & \multicolumn{1}{c|}{25.38} & \color{green}2.50 & 38.16 & \multicolumn{1}{c|}{39.24} & \color{green}1.08 & 31.78 & \multicolumn{1}{c|}{32.14} & \color{green}0.36 & 43.20 & \multicolumn{1}{c|}{44.78} & \color{green}1.58\ \ \ \\
\multicolumn{1}{c|}{} & Pass@10 & 40.00 & \multicolumn{1}{c|}{48.60} & \color{green}8.60 & 64.20 & \multicolumn{1}{c|}{72.40} & \color{green}8.20 & 53.00 & \multicolumn{1}{c|}{57.40} & \color{green}4.40 & 60.80 & \multicolumn{1}{c|}{69.00} & \color{green}8.20\ \ \ \\ \midrule
\multicolumn{1}{c|}{\multirow{2}{*}{MBPP+}} & Pass@1 & 26.50 & \multicolumn{1}{c|}{29.60} & \color{green}3.10 & 47.00 & \multicolumn{1}{c|}{47.90} & \color{green}0.90 & 40.30 & \multicolumn{1}{c|}{41.30} & \color{green}1.00 & 56.00 & \multicolumn{1}{c|}{54.10} & \color{red}1.90\ \ \ \\
\multicolumn{1}{c|}{} & Pass@10 & 39.00 & \multicolumn{1}{c|}{51.00} & \color{green}12.00 & 70.00 & \multicolumn{1}{c|}{68.00} & \color{red}2.00 & 62.00 & \multicolumn{1}{c|}{63.00} & \color{green}1.00 & 65.00 & \multicolumn{1}{c|}{65.00} & 0.00\ \ \ \\\bottomrule
\end{tabular}}
\vspace{-2mm}
\end{table*}

\begin{table*}[]
\caption{Problem-solving performance (Pass@1 and Pass@10, in \%) across three datasets for models tuned with  $\mathcal{R}_s$ and $\mathcal{R}_\zeta$. $\Delta = |\mathcal{R}_\zeta - \mathcal{R}_s|$, where {\color{green} green} indicates that $\mathcal{R}_\zeta$ exceeds that of $\mathcal{R}_s$, while {\color{red} red} indicates the opposite.}
\label{tab:rl}
\vspace{-2mm}
\centering
\resizebox{\textwidth}{!}{
\begin{tabular}{@{}cccccccccccccc@{}}
\toprule
\multicolumn{2}{c|}{Model} & \multicolumn{3}{c|}{Llama-3.2-1B} & \multicolumn{3}{c|}{Llama-3.1-8B} & \multicolumn{3}{c|}{Qwen2.5-1.5B} & \multicolumn{3}{c}{Qwen2.5-7B} \\ \midrule
\multicolumn{1}{c|}{Dataset} & \multicolumn{1}{c|}{\ \ \ \ Metric\ \ \ \ } & \ \ \ \ $\mathcal{R}_s$ \ \ \ \ & \multicolumn{1}{c|}{\ \ \ $\mathcal{R}_\zeta$\ \ \ \ \ } & \multicolumn{1}{c|}{\ \ \ \ $\Delta$\ \ \ \ } & \ \ \ \ $\mathcal{R}_s$ \ \ \ \ & \multicolumn{1}{c|}{\ \ \ $\mathcal{R}_\zeta$\ \ \ \ \ } & \multicolumn{1}{c|}{\ \ \ \ $\Delta$\ \ \ \ } & \ \ \ \ $\mathcal{R}_s$ \ \ \ \ & \multicolumn{1}{c|}{\ \ \ $\mathcal{R}_\zeta$\ \ \ \ \ } & \multicolumn{1}{c|}{\ \ \ \ $\Delta$\ \ \ \ } & \ \ \ \ $\mathcal{R}_s$ \ \ \ \ & \multicolumn{1}{c|}{\ \ \ $\mathcal{R}_\zeta$\ \ \ \ \ } & \ \ \ \ $\Delta$ \ \ \ \ \\ \midrule
\multicolumn{14}{c}{$\mathcal{D}^-_{\mathrm{SFT}}$} \\ \midrule
\multicolumn{1}{c|}{\multirow{2}{*}{Maze}} & \multicolumn{1}{c|}{Pass@1} & 25.60 & \multicolumn{1}{c|}{26.86} & \multicolumn{1}{c|}{\color{green}1.26} & 31.80 & \multicolumn{1}{c|}{32.44} & \multicolumn{1}{c|}{\color{green}0.64} & 29.58 & \multicolumn{1}{c|}{31.34} & \multicolumn{1}{c|}{\color{green}1.76} & 27.28 & \multicolumn{1}{c|}{26.74} &\color{red} 0.54 \\
\multicolumn{1}{c|}{} & \multicolumn{1}{c|}{Pass@10} & 35.40 & \multicolumn{1}{c|}{40.80} & \multicolumn{1}{c|}{\color{green}5.40} & 39.00 & \multicolumn{1}{c|}{46.40} & \multicolumn{1}{c|}{\color{green}7.40} & 37.40 & \multicolumn{1}{c|}{41.80} & \multicolumn{1}{c|}{\color{green}4.40} & 36.20 & \multicolumn{1}{c|}{40.20} & \color{green}4.00 \\ \midrule
\multicolumn{1}{c|}{\multirow{2}{*}{Math-500}} & \multicolumn{1}{c|}{Pass@1} & 24.52 & \multicolumn{1}{c|}{25.52} & \multicolumn{1}{c|}{\color{green}1.00} & 41.12 & \multicolumn{1}{c|}{41.50} & \multicolumn{1}{c|}{\color{green}0.38} & 35.48 & \multicolumn{1}{c|}{36.04} & \multicolumn{1}{c|}{\color{green}0.56} & 46.30 & \multicolumn{1}{c|}{46.26} & \color{red}0.04 \\
\multicolumn{1}{c|}{} & \multicolumn{1}{c|}{Pass@10} & 41.60 & \multicolumn{1}{c|}{41.20} & \multicolumn{1}{c|}{\color{red}0.40} & 68.00 & \multicolumn{1}{c|}{68.20} & \multicolumn{1}{c|}{\color{green}0.20} & 54.20 & \multicolumn{1}{c|}{58.80} & \multicolumn{1}{c|}{\color{green}4.60} & 66.40 & \multicolumn{1}{c|}{68.20} & \color{green}1.80 \\ \midrule
\multicolumn{1}{c|}{\multirow{2}{*}{MBPP+}} & \multicolumn{1}{c|}{Pass@1} & 33.40 & \multicolumn{1}{c|}{33.40} & \multicolumn{1}{c|}{0.00} & 49.40 & \multicolumn{1}{c|}{48.60} & \multicolumn{1}{c|}{\color{red}0.80} & 48.10 & \multicolumn{1}{c|}{47.70} & \multicolumn{1}{c|}{\color{red}0.40} & 59.80 & \multicolumn{1}{c|}{60.30} & \color{green}0.50 \\
\multicolumn{1}{c|}{} & \multicolumn{1}{c|}{Pass@10} & 44.00 & \multicolumn{1}{c|}{50.00} & \multicolumn{1}{c|}{\color{green}6.00} & 59.00 & \multicolumn{1}{c|}{63.00} & \multicolumn{1}{c|}{\color{green}4.00} & 59.00 & \multicolumn{1}{c|}{63.00} & \multicolumn{1}{c|}{\color{green}4.00} & 68.00 & \multicolumn{1}{c|}{69.00} & \color{green}1.00 \\ \midrule
\multicolumn{14}{c}{$\mathcal{D}^+_{\mathrm{SFT}}$} \\ \midrule
\multicolumn{1}{c|}{\multirow{2}{*}{Maze}} & \multicolumn{1}{c|}{Pass@1} & 32.60 & \multicolumn{1}{c|}{30.94} & \multicolumn{1}{c|}{\color{red}1.66} & 37.82 & \multicolumn{1}{c|}{34.26} & \multicolumn{1}{c|}{\color{red}3.56} & 31.06 & \multicolumn{1}{c|}{30.66} & \multicolumn{1}{c|}{\color{red}0.40} & 26.64 & \multicolumn{1}{c|}{27.34} & \color{green}0.70 \\
\multicolumn{1}{c|}{} & \multicolumn{1}{c|}{Pass@10} & 43.20 & \multicolumn{1}{c|}{47.60} & \multicolumn{1}{c|}{\color{green}4.40} & 43.00 & \multicolumn{1}{c|}{54.40} & \multicolumn{1}{c|}{\color{green}11.40} & 43.40 & \multicolumn{1}{c|}{46.40} & \multicolumn{1}{c|}{\color{green}3.00} & 33.00 & \multicolumn{1}{c|}{43.80} & \color{green}10.80 \\ \midrule
\multicolumn{1}{c|}{\multirow{2}{*}{Math-500}} & \multicolumn{1}{c|}{Pass@1} & 26.92 & \multicolumn{1}{c|}{27.44} & \multicolumn{1}{c|}{\color{green}0.52} & 43.58 & \multicolumn{1}{c|}{44.26} & \multicolumn{1}{c|}{\color{green}0.68} & 34.42 & \multicolumn{1}{c|}{34.78} & \multicolumn{1}{c|}{\color{green}0.36} & 47.50 & \multicolumn{1}{c|}{50.22} & \color{green}2.72 \\
\multicolumn{1}{c|}{} & \multicolumn{1}{c|}{Pass@10} & 50.20 & \multicolumn{1}{c|}{49.20} & \multicolumn{1}{c|}{\color{red}1.00} & 71.80 & \multicolumn{1}{c|}{73.48} & \multicolumn{1}{c|}{\color{green}1.68} & 57.60 & \multicolumn{1}{c|}{62.40} & \multicolumn{1}{c|}{\color{green}4.80} & 69.00 & \multicolumn{1}{c|}{70.80} & \color{green}1.80 \\ \midrule
\multicolumn{1}{c|}{\multirow{2}{*}{MBPP+}} & \multicolumn{1}{c|}{Pass@1} & 37.90 & \multicolumn{1}{c|}{38.60} & \multicolumn{1}{c|}{\color{green}0.70} & 46.90 & \multicolumn{1}{c|}{50.10} & \multicolumn{1}{c|}{\color{green}3.20} & 45.90 & \multicolumn{1}{c|}{47.60} & \multicolumn{1}{c|}{\color{green}1.70} & 54.00 & \multicolumn{1}{c|}{55.00} & \color{green}1.00 \\
\multicolumn{1}{c|}{} & \multicolumn{1}{c|}{Pass@10} & 58.00 & \multicolumn{1}{c|}{58.80} & \multicolumn{1}{c|}{\color{green}0.80} & 63.00 & \multicolumn{1}{c|}{65.00} & \multicolumn{1}{c|}{\color{green}2.00} & 64.00 & \multicolumn{1}{c|}{70.00} & \multicolumn{1}{c|}{\color{green}6.00} & 63.00 & \multicolumn{1}{c|}{66.00} & \color{green}3.00 \\ \bottomrule
\end{tabular}}
\vspace{-2mm}
\end{table*}

\subsection{Main Results}
\paragraph{Dataset Divergence Metric.}

Tab.~\ref{tab:sft} reports problem-solving performance and solution divergence across the three datasets for models fine-tuned on low- and high-divergence training samples. Models trained on $\mathcal{D}_{\text{SFT}}^+$ outperform those trained on $\mathcal{D}_{\text{SFT}}^-$ in 8 out of 12 cases for Pass@1 ($\text{mean}(\Delta)=0.65\%$). The advantage is even clearer for Pass@10, where $\mathcal{D}_{\text{SFT}}^+$ yields higher performance in 10 out of 12 cases ($\text{mean}(\Delta)=6.2\%$). These results underscore the strong influence of solution divergence in training data on SFT model performance, supporting divergence-based sample selection as an effective strategy for improving problem-solving ability. An exception is observed on the MBPP+ dataset, where performance differences between low- and high-divergence training are minimal. Examination of the training process indicates that the limited size of MBPP+ leads to rapid overfitting and early stopping, thereby reducing the benefits of divergence-based sample selection. Finally, as some values of $\Delta$ are small, we repeat the experiment over small sized model for multiple times and report their statical significancy in Appx.~\ref{sec:repeat}.

\paragraph{Solution Divergence Fused Reward.}

We now present the RL results in Tab.~\ref{tab:rl}. When applying RL to models initialized from $\mathcal{D}_{\text{SFT}}^-$, the divergence-fused reward $\mathcal{R}_\zeta$ outperforms the binary success reward $\mathcal{R}_s$ on Pass@1 in 7 out of 12 cases, with an average improvement of 0.34\%. More importantly, on Pass@10, models trained with $\mathcal{R}_\zeta$ achieve overwhelming advantages in 11 out of 12 cases, with an average improvement of 3.12\% over $\mathcal{R}_s$. For models initialized from $\mathcal{D}_{\text{SFT}}^+$, we observe a consistent trend. Specifically, $\mathcal{R}_\zeta$ improves Pass@1 performance in 9 out of 12 cases, and the average performance outperforms $\mathcal{R}_s$ by 0.50\%. For Pass@10, $\mathcal{R}_\zeta$ also maintains its advantage, outperforming $\mathcal{R}_s$ in 11 out of 12 cases with an average gain of 4.06\%. These findings provide strong evidence that the divergence-fused reward is effective in enhancing the problem-solving ability of LLMs. Notably, in several cases $\mathcal{R}_\zeta$ yields lower Pass@1 performance but significantly better Pass@10 results. This suggests that $\mathcal{R}_\zeta$ encourages broader exploration of solution space, which helps models discover diverse solutions to new problems, rather than relying solely on rigid, known solutions. Together, these results confirm the effectiveness of incorporating solution divergence into the reward function. Additionally, we run multiple experiments using smaller-sized models and report their statistical significance in Appx.~\ref{sec:repeat}. 

\paragraph{Generation Behaviors.}
Beyond performance metrics, we further compare generation behavior statistics between models trained with and without solution-divergence-based interventions, including generation length and average token entropy. We observe that encouraging solution divergence increases generation entropy without substantially increasing generation length, suggesting broader and more effective exploration of the solution space rather than the production of meaningless tokens; additional analyses are provided in Appx.~\ref{appx:behavior}.

\subsection{Ablation Studies}
\label{sec:ablation}

In addition to the main results, we conduct ablation studies on the Maze dataset, where task performance is highly sensitive to variations in solution divergence. This sensitivity makes Maze a particularly suitable testbed for uncovering deeper insights into how solution divergence influences model behavior across different settings. Moreover, as Maze is a novel problem-solving task introduced in this work, using it for case analyses helps avoid potential data-contamination issues that may affect the other two public benchmark datasets. Finally, we conduct temperature-tuning studies and evaluate different proxy metrics for pairwise solution divergence using two small-scale models across all datasets, disentangling the benefits of the solution-divergence-fused reward from those arising simply from increased sampling diversity.

\paragraph{Data Size Influence.}

\begin{figure}[!btph]
     \centering
     \begin{subfigure}[b]{0.475\textwidth}
         \centering
         \includegraphics[width=\textwidth]{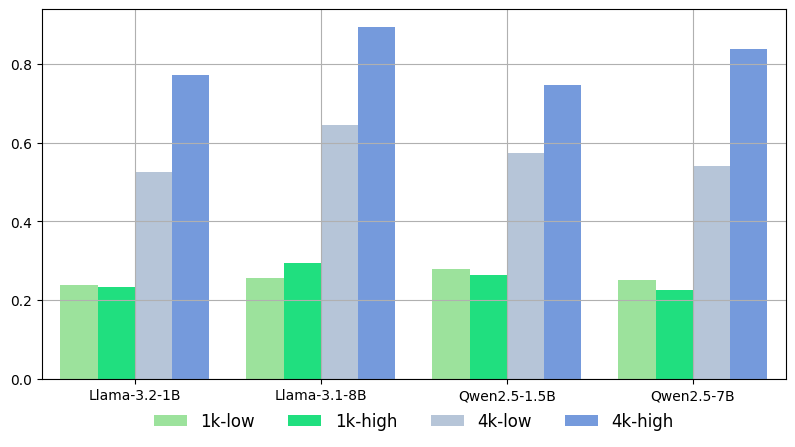}
         \vspace{-5mm}
         \caption{Pass@1}
         \label{fig:1k4k_pass1}
     \end{subfigure}
     \begin{subfigure}[b]{0.475\textwidth}
         \centering
         \includegraphics[width=\textwidth]{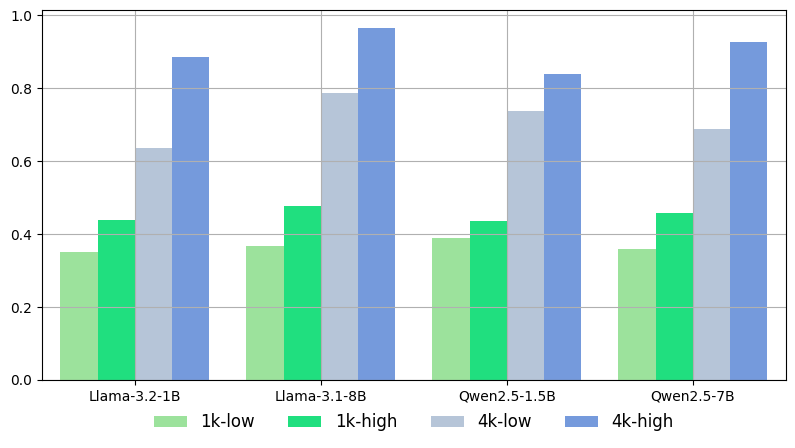}
         \vspace{-5mm}
         \caption{Pass@10}
         \label{fig:1k4k_pass10}
     \end{subfigure}
     \vspace{-2mm}
     \caption{Performance of models trained on $\mathcal{D}_{\text{SFT}}^-$ (1k-low), $\mathcal{D}_{\text{SFT}}^{--}$ (4k-low), $\mathcal{D}_{\text{SFT}}^+$ (1k-high), and $\mathcal{D}_{\text{SFT}}^{++}$ (4k-high) for solving problems in Maze.}
     \label{fig:exp_1k4k}
     \vspace{-5mm}
\end{figure}

We investigate the influence of SFT training data size on the relationship between solution divergence and LLM problem-solving performance. Specifically, we expand the number of unique questions in $\mathcal{D}_{\text{SFT}}^+$ and $\mathcal{D}_{\text{SFT}}^-$ from 250 to 1,000, thereby increasing the dataset sizes from 1,000 to 4,000. This yields two new datasets: $\mathcal{D}_{\text{SFT}}^{++}$ and $\mathcal{D}_{\text{SFT}}^{--}$. Fig.~\ref{fig:exp_1k4k} reports model performance across all four SFT datasets. The results show that training with more samples improves overall performance, while also widening the gap between low- and high-divergence datasets, further highlighting the benefit of incorporating solution divergence in SFT training.

\paragraph{Performance and Divergence Balanced Reward.}

\begin{figure}[!btph]
     \centering
     \begin{subfigure}[b]{0.475\textwidth}
         \centering
         \includegraphics[width=\textwidth]{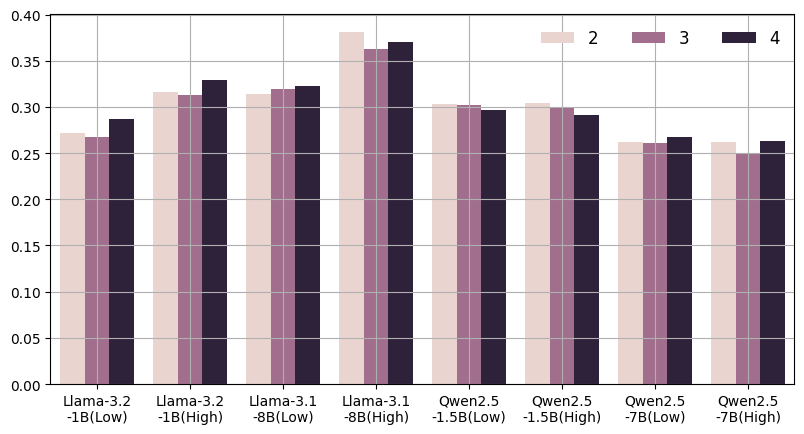}
         \vspace{-5mm}
         \caption{Pass@1}
         \label{fig:alpha_pass1}
     \end{subfigure}
     \begin{subfigure}[b]{0.475\textwidth}
         \centering
         \includegraphics[width=\textwidth]{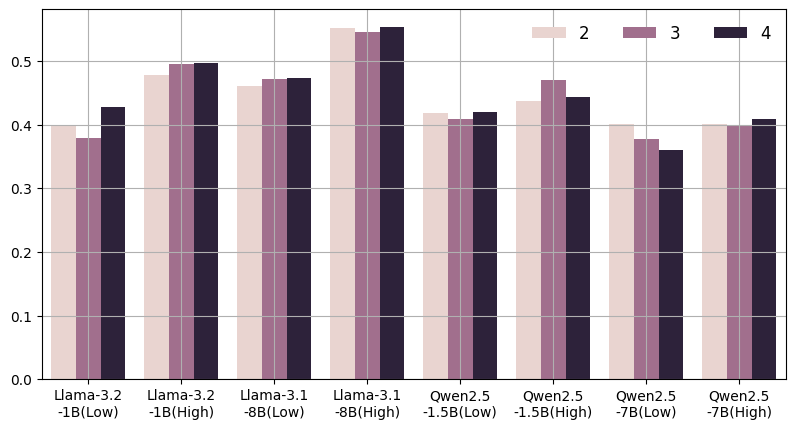}
         \vspace{-5mm}
         \caption{Pass@10}
         \label{fig:alpha_pass10}
     \end{subfigure}
     \vspace{-2mm}
     \caption{Performance of models trained on $\alpha=2,3,4$ for solving problems in Maze.}
     \label{fig:exp_234}
     \vspace{-5mm}
\end{figure}

In \S~\ref{sec:rl_reward}, we introduced the hyper-parameter $\alpha$ to balance solution divergence and problem-solving performance during RL training. We experiment with $\alpha \in \{2,3,4\}$, which scales the divergence term to be inversely proportional, constant, or directly proportional to the success rate. As shown in Fig.~\ref{fig:exp_234}, $\alpha=4$ yields the best performance in 5 out of 8 Pass@1 cases and 6 out of 8 Pass@10 cases. This suggests that it is generally beneficial to limit the influence of divergence when problem-solving performance is low, and to amplify it as performance improves. This finding aligns with cognitive science theory~\citep{siegler1998emerging}, which emphasizes first developing a correct strategy and only then expanding to diverse alternatives.

\begin{table*}[]
\caption{Problem-solving performance (Pass@1 and Pass@10, in \%) on Maze using different generating temperatures. Results are shown for models fine-tuned with $\mathcal{D}_{\text{SFT}}^-$, $\mathcal{D}_{\text{SFT}}^+$, $\mathcal{R}_s$ and $\mathcal{R}_{\zeta}$.}
\label{tab:temperature}
\vspace{-2mm}
\centering
\resizebox{\textwidth}{!}{
\begin{tabular}{@{}cc|cccccc|cccccc@{}}
\toprule
\multicolumn{2}{c|}{Model} & \multicolumn{6}{c|}{Llama-3.2-1B} & \multicolumn{6}{c}{Qwen2.5-1.5B} \\ \midrule
\multicolumn{2}{c|}{\ \ \ Method\ \ \ } & \multicolumn{2}{c|}{\ \ \ \ \ $\mathcal{D}^-_{\text{SFT}}$\ \ \ \ \ } & \multicolumn{1}{c|}{\ \ \ $\mathcal{D}^+_{\text{SFT}}$\ \ \ } & \multicolumn{2}{c|}{\ \ \ $\mathcal{R}_s$\ \ \ } &\ \ \ \ \ $\mathcal{R}_\zeta$ \ \ \ \ \ & \multicolumn{2}{c|}{\ \ \ \ \ $\mathcal{D}^-_{\text{SFT}}$\ \ \ \ \ } & \multicolumn{1}{c|}{\ \ \ $\mathcal{D}^+_{\text{SFT}}$\ \ \ } & \multicolumn{2}{c|}{\ \ \ \ \ $\mathcal{R}_s$\ \ \ \ \ } &\ \ \ \ \ $\mathcal{R}_\zeta$\ \ \ \ \ \\ \midrule
\multicolumn{2}{c|}{Temperature} & 0.9 & \multicolumn{1}{c|}{1.2} & \multicolumn{1}{c|}{0.6} & 0.9 & \multicolumn{1}{c|}{1.2} & 0.6 & 0.9 & \multicolumn{1}{c|}{1.2} & \multicolumn{1}{c|}{0.7} & 0.9 & \multicolumn{1}{c|}{1.2} & 0.7 \\ \midrule
\multicolumn{1}{c|}{\multirow{3}{*}{Maze}} & \ \ \ \ Pass@1 \ \ \ \ & \ \ 22.54 \ \ & \multicolumn{1}{c|}{\ \ 21.08\ \ } & \multicolumn{1}{c|}{\ \ 23.40\ \ } & \ \ 25.10 \ \ & \multicolumn{1}{c|}{\ \ 23.58\ \ } & \ \ 26.86 \ \ & \ \ 27.45 \ \ & \multicolumn{1}{c|}{\ \ 27.04\ \ } & \multicolumn{1}{c|}{\ \ 26.48\ \ } & \ \ 29.09 \ \ & \multicolumn{1}{c|}{\ \ 28.85\ \ } & \ \ 31.34 \ \ \ \\
\multicolumn{1}{c|}{} & Pass@10 & 40.00 & \multicolumn{1}{c|}{41.66} & \multicolumn{1}{c|}{43.80} & 38.20 & \multicolumn{1}{c|}{41.68} & 40.80 & 39.88 & \multicolumn{1}{c|}{42.84} & \multicolumn{1}{c|}{43.60} & 38.32 & \multicolumn{1}{c|}{41.12} & 41.80 \ \\ \cmidrule(l){2-14} 
\multicolumn{1}{c|}{} & Average & 31.27 & \multicolumn{1}{c|}{31.37} & \multicolumn{1}{c|}{33.60} & 31.65 & \multicolumn{1}{c|}{32.63} & 33.83 & 33.67 & \multicolumn{1}{c|}{34.94} & \multicolumn{1}{c|}{35.04} & 33.71 & \multicolumn{1}{c|}{34.99} & 36.57 \ \\ \midrule
\multicolumn{1}{c|}{\multirow{3}{*}{Math-500}} & Pass@1 & 22.66 & \multicolumn{1}{c|}{22.20} & \multicolumn{1}{c|}{25.38} & 23.31 & \multicolumn{1}{c|}{21.34} & 25.52 & 31.75 & \multicolumn{1}{c|}{31.10} & \multicolumn{1}{c|}{32.14} & 35.69 & \multicolumn{1}{c|}{35.28} & 36.04 \ \\
\multicolumn{1}{c|}{} & Pass@10 & 41.32 & \multicolumn{1}{c|}{42.12} & \multicolumn{1}{c|}{48.60} & 43.30 & \multicolumn{1}{c|}{43.60} & 42.20 & 54.56 & \multicolumn{1}{c|}{55.84} & \multicolumn{1}{c|}{57.40} & 56.60 & \multicolumn{1}{c|}{59.00} & 58.80 \ \\ \cmidrule(l){2-14} 
\multicolumn{1}{c|}{} & Average & 31.99 & \multicolumn{1}{c|}{32.16} & \multicolumn{1}{c|}{36.99} & 33.31 & \multicolumn{1}{c|}{32.47} & 33.86 & 43.16 & \multicolumn{1}{c|}{43.47} & \multicolumn{1}{c|}{44.77} & 46.15 & \multicolumn{1}{c|}{47.14} & 47.42 \ \\ \midrule
\multicolumn{1}{c|}{\multirow{3}{*}{MBPP+}} & Pass@1 & 24.02 & \multicolumn{1}{c|}{21.42} & \multicolumn{1}{c|}{29.60} & 33.94 & \multicolumn{1}{c|}{32.00} & 33.40 & 40.12 & \multicolumn{1}{c|}{34.80} & \multicolumn{1}{c|}{41.30} & 44.23 & \multicolumn{1}{c|}{27.02} & 47.70 \ \\
\multicolumn{1}{c|}{} & Pass@10 & 41.00 & \multicolumn{1}{c|}{53.00} & \multicolumn{1}{c|}{51.00} & 42.92 & \multicolumn{1}{c|}{49.37} & 50.00 & 62.40 & \multicolumn{1}{c|}{65.80} & \multicolumn{1}{c|}{63.00} & 64.38 & \multicolumn{1}{c|}{64.38} & 63.00 \ \\ \cmidrule(l){2-14} 
\multicolumn{1}{c|}{} & Average & 32.51 & \multicolumn{1}{c|}{37.21} & \multicolumn{1}{c|}{40.30} & 38.43 & \multicolumn{1}{c|}{40.69} & 41.70 & 51.26 & \multicolumn{1}{c|}{50.30} & \multicolumn{1}{c|}{52.15} & 54.31 & \multicolumn{1}{c|}{45.70} & 55.35 \ \\ \bottomrule
\end{tabular}}
\vspace{-2mm}
\end{table*}

\paragraph{Generation Temperature Tuning.}
Tuning the generation temperature is a common technique for adjusting a model’s output style without retraining. In general, higher temperatures encourage more diverse and creative generations, which can lead to improvements in Pass@10 on problem-solving benchmarks. To compare the effect of temperature tuning with our proposed solution-divergence–based training, we conduct the following experiments. For the SFT stage, we take the model trained on $\mathcal{D}^-_{\text{SFT}}$ and generate solutions using two additional temperatures: 0.9, and 1.2. We compare these results against the model trained on $\mathcal{D}^+_{\text{SFT}}$, evaluated with the default temperature. For the RL stage, we similarly evaluate the model trained with $\mathcal{R}_s$ at different temperatures and compare it to the model trained with $\mathcal{R}_{\zeta}$ under the default temperature. Results are reported in Tab.~\ref{tab:temperature}. Because both Pass@1 and Pass@10 are important, we additionally report their average. From the table, we observe that increasing the temperature improves the Pass@10 of models trained on $\mathcal{D}^-_{\text{SFT}}$ and $\mathcal{R}_s$, but typically reduces Pass@1. When comparing the averages of the two metrics, models trained with $\mathcal{D}^+_{\text{SFT}}$ and $\mathcal{R}_{\zeta}$ consistently maintain the best performance. These results suggest that solution-divergence–based training provides more robust overall gains than temperature tuning alone. In addition to the Maze, we also experiment result experiments with both the MBPP and Math datasets and the consistent observation can be observed.

\paragraph{Pair-wise Divergence Proxy Tuning.}
\label{sec:embedding}

We investigate how different pairwise divergence proxy metrics influence our findings. Specifically, we consider three types of approximation methods: edit distance $d^{(e)}$, embedding cosine distance $d^{(c)}$, and natural language inference (NLI) score $d^{(i)}$. For the embedding-based proxy, we use the Sentence-BERT model all-MiniLM-L6-v2~\citep{reimers2019sentence} to encode each solution into a vector representation and compute divergence using cosine distance. For the NLI-based proxy, we employ bart-large-mnli~\citep{lewis2020bart} to estimate pairwise entailment scores between solutions. We focus on evaluating the influence of these proxy metrics during the RL training phase using Qwen2.5-1.5B. As shown in Tab.~\ref{tab:embedding}, different tasks exhibit different preferences for divergence proxy metrics. For the Maze dataset, $d^{(e)}$ achieves the best performance, while for both Math and MBPP, $d^{(c)}$ performs best. This phenomenon can be explained by the nature of the solution formats. Solutions in Maze are primarily structured operation sequences with limited semantic meaning, making the simpler edit-distance-based proxy more appropriate. In contrast, solutions in Math and MBPP contain richer semantic information and more complex reasoning structures, allowing embedding-based metrics to better capture meaningful divergence between solutions. As for the NLI-based proxy, its performance is consistently weaker than the other two methods. One possible explanation is that the original NLI model is primarily designed for sentence-level entailment prediction rather than fine-grained solution divergence estimation, making it less suitable when directly applied to this setting.

\begin{table}[]
\caption{Performance (Pass@1 and Pass@10, in \%) using different pair-wised solution divergence proxy ways. Results are shown for models fine-tuned with the divergence introduced reward function $\mathcal{R}_{\zeta}$.}
\label{tab:embedding}
\centering
\resizebox{.475\textwidth}{!}{
\begin{tabular}{@{}c|c|ccc@{}}
\toprule
\ \ \ \ Dataset \ \ \ \ &\ \ \ \ Metric \ \ \ \ & \ \ \ \ \ \ \ $d^{(e)}$\ \ \ \ \ &\ \ \ \ \ $d^{(c)}$\ \ \ \ \ &\ \ \ \ \ $d^{(i)}$ \ \ \ \ \ \ \\ \midrule
\multirow{2}{*}{Maze} & Pass@1 & 31.04 & 23.76 & 26.66 \\
 & Pass@10 & 42.00 & 36.20 & 33.20 \\ \midrule
\multirow{2}{*}{Math-500} & Pass@1 & 39.52 & 44.68 & 42.22 \\
 & Pass@10 & 66.40 & 70.40 & 58.00 \\ \midrule
\multirow{2}{*}{MBPP+} & Pass@1 & 50.60 & 54.20 & 54.20 \\
 & Pass@10 & 62.00 & 66.00 & 57.00 \\ \bottomrule
\end{tabular}}
\vspace{-5mm}
\end{table}

%% file: conclusion.tex
In this paper, we investigate an underexplored direction for enhancing the problem-solving performance of LLMs: solution divergence, defined as the presence of multiple viable solutions to a single problem. Our preliminary study empirically demonstrates a positive relationship between solution divergence and model performance. Building on this insight, we introduce two methods, dataset divergence metric and divergence-fused reward, to augment existing SFT and RL algorithms. Comprehensive experiments across three representative problem-solving tasks in the logical reasoning, mathematics, and programming domains confirm the effectiveness of leveraging solution divergence to improve LLM performance. These findings highlight the potential of solution divergence as a valuable training signal and open new avenues for future research on harnessing diversity in solutions to strengthen LLM problem-solving capabilities.

%% file: appendix.tex
\section{Preliminary Study Dataset Details}
\label{appx:pre_data}

We conduct our preliminary study on the relationship between LLMs’ solution divergence ($\zeta_\pi$) and problem-solving performance (Pass@1) across three datasets: Math-500, MBPP+, and Maze.

\paragraph{Math-500.} This dataset is a high-quality subset of Math, a widely used benchmark for evaluating LLMs’ mathematical problem-solving ability. Unlike the full Math dataset, which includes standard train/validation/test splits, Math-500 contains 500 correctness-verified questions sampled from the original test split. The dataset spans seven categories and provides ground-truth answers, with most questions also annotated with difficulty levels (1–5). For efficiency and to control API costs when querying closed-source models, we randomly sampled 100 questions from Math-500 for our experiments.

\paragraph{MBPP+.} MBPP+ is a refined version of the MBPP benchmark, designed to evaluate LLMs’ ability to solve Python programming tasks. Each problem specifies a function signature and requires a correct implementation that passes the associated test cases. Compared to MBPP, MBPP+ improves test robustness by expanding coverage of edge cases; a small number of ambiguous problems from MBPP were removed. In total, MBPP+ contains 378 problems, aligned with the original MBPP train/validation/test split. For consistency with Math-500, we randomly sampled 100 test problems to form our preliminary programming dataset. Both Math-500 and MBPP+ are accessed via the Hugging Face datasets library\footnote{\url{https://huggingface.co/docs/datasets/index}}.

\paragraph{Maze.} Maze is a new logical reasoning dataset introduced in this paper. Each problem asks an LLM to find a viable path from a fixed start point (0,0) to a random goal point within a $10\times 10$ grid. To increase difficulty, we add a blocking set $\mathcal{B}$, where $|\mathcal{B}|<100$, with blocked coordinates sampled uniformly at random. Each problem instance has a distinct blocking configuration. Following the setup for Math-500 and MBPP+, we generated 100 Maze problems for the preliminary study (see Tab.~\ref{tab:pre_example} for an illustration).

\paragraph{Verification.} To verify solution correctness across tasks, we combine existing open-source verification tools with additional rules tailored to each dataset. For Math-500, we use the open-source math\_verify package\footnote{\url{https://github.com/huggingface/Math-Verify}}. This tool processes full solution strings, extracts the marked answer, and handles common numerical-equivalence cases with high accuracy. Since our prompts instruct the model to generate step-by-step solutions, we extract the final answer from the last step and pass it, along with the ground-truth answer, to the verifier. For MBPP+, we execute each generated function against the dataset’s official Python test cases. We cap the runtime of each test at 200 ms; if execution raises an error or exceeds the time limit, the solution is marked incorrect. For Maze, we reconstruct the complete movement trajectory from the generated solution. A solution is judged correct if the final position reaches the goal and all intermediate positions avoid any blocked coordinates in $\mathcal{B}$.

\section{Preliminary Study Models}
\label{appx:pre_model}

To ensure the robustness of our conclusion, we comprehensively select 17 representative LLMs from both open- and close-sourced ones. Below, we list them by their series names as follows: o1-mini, o1-mini, gpt-4o, gpt-4o-mini, claude-3-7-sonnet-20250219, claude-3-5-sonnet-20241022, gemini-1.5-pro-002, gemini-1.5-flash-002, gemini-2.0-flash-001, Llama-3.1-70B-Instruct, Llama-3.1-8B-Instruct, Qwen2.5-72B-Instruct, Qwen2.5-7B-Instruct, DeepSeek-R1-Distill-Qwen-32B, DeepSeek-R1-Distill-Qwen-7B, DeepSeek-R1-Distill-Llama-70B, DeepSeek-R1-Distill-Llama-8B. For all the close-sourced ones, we implement based on the huggingface Transformer package\footnote{\url{https://huggingface.co/docs/transformers/en/index}} and we use the default generation configuration for each model inference. For the open-sourced LLMs. we send request to official API endpoints for the results.

\section{Divergence Reward Simplification}
\label{appx:reward_simplify}

In \S~\ref{sec:rl_reward}, we presented the simplified form of the reward function in Eq.~\ref{eq:reward}. For completeness, we provide the full derivation here. We start from the original formulation:
\begin{equation}
    \mathcal{R}_n = \sum_{s_i \in \mathcal{S}}
    \mathcal{R}_\zeta(s_i, \mathcal{S}) \nonumber
\end{equation}

\noindent Substituting Eq.~\ref{eq:reward_def} into Eq.~\ref{eq:reward}, we have
\begin{align}
    \label{eq:simple_2}
    &\mathcal{R}_n
    = \sum_{s_i \in \mathcal{S}_c}
    \left(\frac{|\mathcal{S}_c|}{|\mathcal{S}|}\right)^\alpha
    \frac{\sum_{s_j \in \mathcal{S}_c}\delta(s_i,s_j)}{|\mathcal{S}_c|}
    - \sum_{s_i \notin \mathcal{S}_c} 1 \nonumber \\
    &= \left(\frac{|\mathcal{S}_c|}{|\mathcal{S}|}\right)^{\alpha-1}\frac{1}{|\mathcal{S}|}
    \sum_{s_i \in \mathcal{S}_c}\sum_{s_j \in \mathcal{S}_c}\delta(s_i,s_j) \\
    &- \big(|\mathcal{S}| - |\mathcal{S}_c|\big).
\end{align}

\noindent Based on the Eq.~\ref{eq:div_def}, we have:

\begin{align*}
    \zeta_{q_n}^{g} &= M - \frac{1}{M}\sum^M_{i=1}\lambda_i = M - \frac{1}{M}\text{tr}(\Lambda) \\
    &= M - \frac{1}{M}\text{tr}(L) = M - \frac{1}{M}\text{tr}(D-A) \\
    &= M - \frac{1}{M}\text{tr}(D) \\
    &= M - \frac{1}{M}\sum_{s_i\in \mathcal{S}}\sum_{s_j\in \mathcal{S}}(1-\delta(s_i,s_j))
\end{align*}

\noindent By definition $|\mathcal{S}|=M$, thus we have:

\begin{align}
\label{eq:simple_1}
    \zeta_{q_n}^g &= |\mathcal{S}| - \frac{1}{|\mathcal{S}|}\sum_{s_i\in \mathcal{S}}\sum_{s_j\in \mathcal{S}}(1-\delta(s_i,s_j)) \nonumber \\
    &= \frac{1}{|\mathcal{S}|}\sum_{s_i\in \mathcal{S}}\sum_{s_j\in \mathcal{S}}\delta(s_i,s_j)
\end{align}

To be noticed, if all the solutions in $\mathcal{S}$ are correct, then we will have $|\mathcal{S}|=|\mathcal{S_c}|$ and the leading term in Eq.~\ref{eq:reward} will always be the constant 1, and the expression won't be influenced by $\alpha$, thus here we only consider about when $|\mathcal{S}_c|<|\mathcal{S}|$. In \S~\ref{sec:pre_setting}, we mention we will conduct the random over-sampling over $\mathcal{S}_c$ to match the size of relation graph $|\mathcal{G}|=|\mathcal{S}|$. Suppose we sample all the solutions in $\mathcal{S}_c$ for $k$ times samples to fulfill the requests, and $k=|\mathcal{S}|/|\mathcal{S}_c|$. We can have: 

\begin{align*}
    &\frac{\sum_{s_i\in \mathcal{S}}\sum_{s_j\in \mathcal{S}}\delta(s_i,s_j)}{\sum_{s_i\in \mathcal{S}_c}\sum_{s_j\in \mathcal{S}_c}\delta(s_i,s_j)} \approx \frac{P(|\mathcal{S}|,2)-|\mathcal{S}_c| P(k,2)}{P(|\mathcal{S}_c|,2)} \\
    &= \frac{|\mathcal{S}|(|\mathcal{S}|-1)-|\mathcal{S}_c|\cdot k(k-1)}{|\mathcal{S}_c|(|\mathcal{S_c}|-1)}\\
    &=\frac{|\mathcal{S}|(|\mathcal{S}|-1)-|\mathcal{S}|\cdot(\frac{|\mathcal{S}|}{|\mathcal{S}_c|}-1)}{|\mathcal{S}_c|(|\mathcal{S}_c|-1)}=\frac{|\mathcal{S}|^2}{|\mathcal{S}_c|^2}
\end{align*}

\noindent where $P(\cdot,\cdot)$ denotes for permutation operation. Plug it back to Eq.~\ref{eq:simple_1}, $\zeta_{q_n}^{g}$ can be expressed as: 

\begin{align*}
    \zeta_{q_n}^{g} \approx \frac{1}{|\mathcal{S}|}\cdot\frac{|\mathcal{S}|^2}{|\mathcal{S}_c|^2}\sum_{s_i\in \mathcal{S}_c}\sum_{s_j\in \mathcal{S}_c}\delta(s_i,s_j)
\end{align*}

\noindent Then, if we do a simply conversion to the Eq.~\ref{eq:simple_2}, we can have:

\begin{align*}
    &\mathcal{R}_n=\left(\frac{|\mathcal{S}_c|}{|\mathcal{S}|}\right)^{\alpha-3} \cdot \frac{1}{|\mathcal{S}|}\cdot\frac{|\mathcal{S}|^2}{|\mathcal{S}_c|^2}\sum_{s_i\in\mathcal{S}_c}\sum_{s_j\in\mathcal{S}_c}\delta(s_i,s_j) \\
    &- (|\mathcal{S}|-|\mathcal{S}_c|) \approx \left(\frac{|\mathcal{S}_c|}{|\mathcal{S}|}\right)^{\alpha-3}\cdot\zeta^g_{q_n}+|\mathcal{S}_c|-|\mathcal{S}|
\end{align*}

\noindent which completes the proof.

\section{Preliminary Study Task Prompts}
\label{appx:pre_prompt}

Below, we present the query prompts for each task. To improve the accuracy of approximating pairwise divergence $\delta_{i,j}$ via string edit distance, we explicitly include output-format instructions in each prompt, requiring LLMs to produce solutions in a standardized style. This design reduces formatting noise and expression redundancy, ensuring cleaner comparisons during our experiments.

\section{Additional Preliminary Study Results}
\label{appx:pre_line}

\begin{figure*}[!btph]
     \centering
     \begin{subfigure}[b]{0.32\textwidth}
         \centering
         \includegraphics[width=\textwidth]{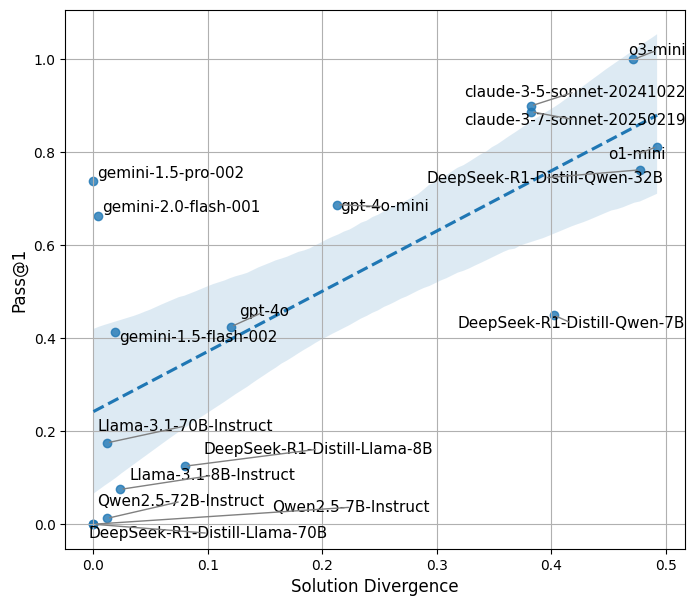}
         \caption{Maze [$\zeta^g_{\pi(e)}$] ($\beta=1.30$)}
         \label{fig:pre_maze_easy}
     \end{subfigure}
     \begin{subfigure}[b]{0.32\textwidth}
         \centering
         \includegraphics[width=\textwidth]{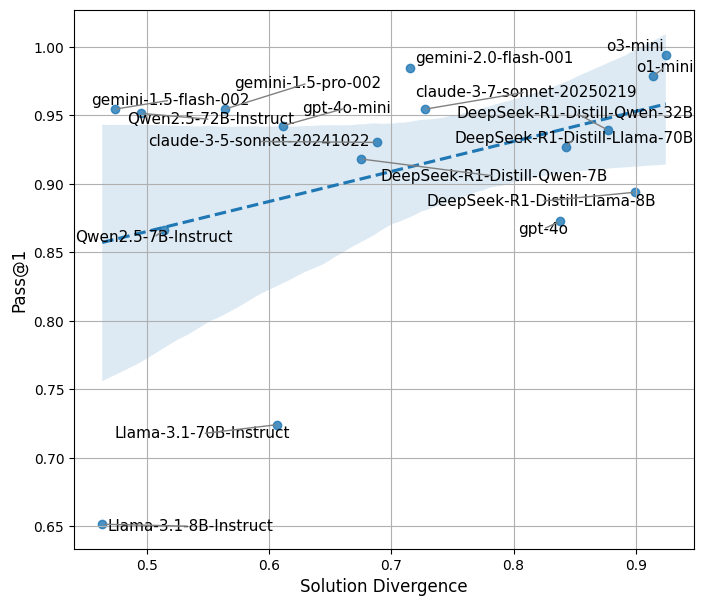}
         \caption{MATH-500 [$\zeta^g_{\pi(e)}$] ($\beta=0.22$)}
         \label{fig:pre_math_easy}
     \end{subfigure}
     \begin{subfigure}[b]{0.32\textwidth}
         \centering
         \includegraphics[width=\textwidth]{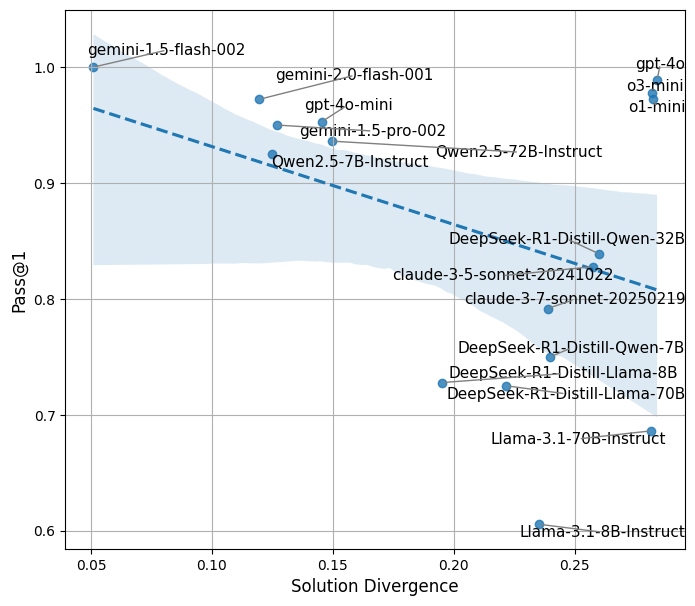}
         \caption{MBPP+ [$\zeta^g_{\pi(e)}$] ($\beta=-0.67$)}
         \label{fig:pre_mbpp_easy}
     \end{subfigure}
     \begin{subfigure}[b]{0.32\textwidth}
         \centering
         \includegraphics[width=\textwidth]{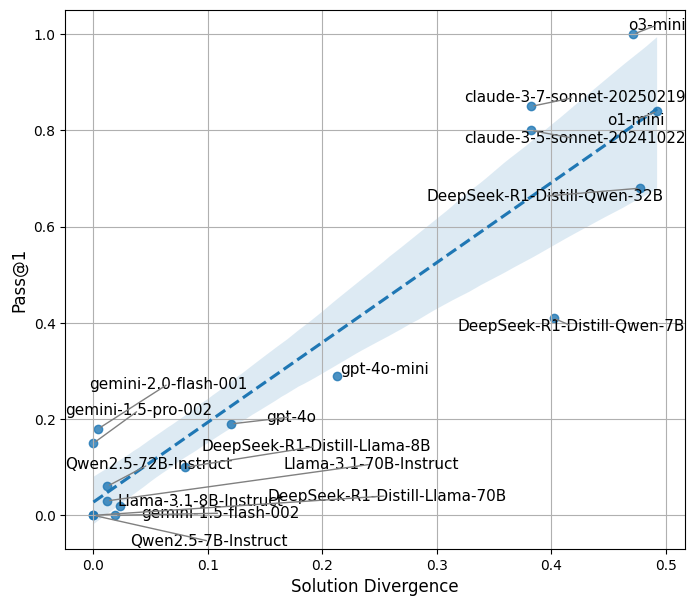}
         \caption{Maze [$\zeta^g_{\pi(m)}$] ($\beta=1.66$)}
         \label{fig:pre_maze_mid}
     \end{subfigure}
     \begin{subfigure}[b]{0.32\textwidth}
         \centering
         \includegraphics[width=\textwidth]{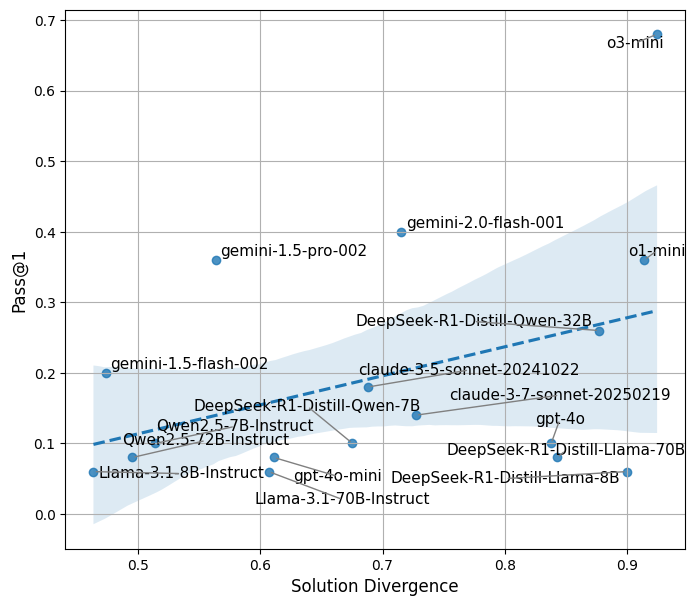}
         \caption{MATH-500 [$\zeta^g_{\pi(m)}$] ($\beta=0.41$)}
         \label{fig:pre_math_mid}
     \end{subfigure}
     \begin{subfigure}[b]{0.32\textwidth}
         \centering
         \includegraphics[width=\textwidth]{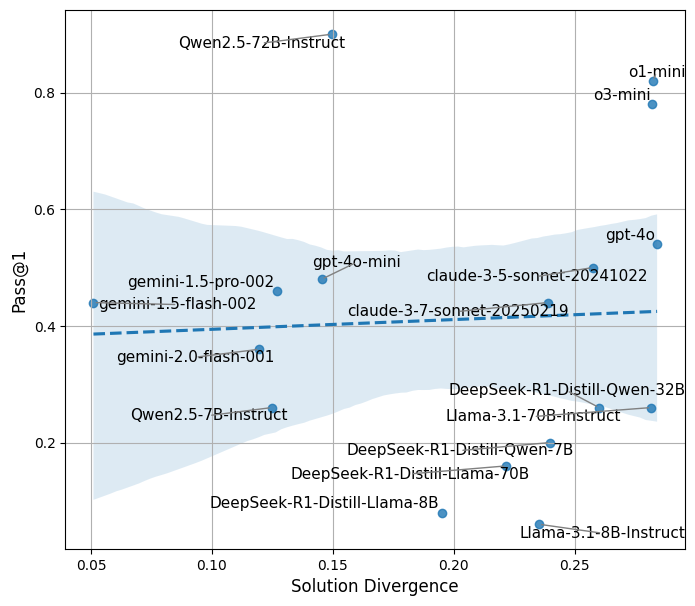}
         \caption{MBPP+ [$\zeta^g_{\pi(m)}$] ($\beta=0.17$)}
         \label{fig:pre_mbpp_mid}
     \end{subfigure}
     \begin{subfigure}[b]{0.32\textwidth}
         \centering
         \includegraphics[width=\textwidth]{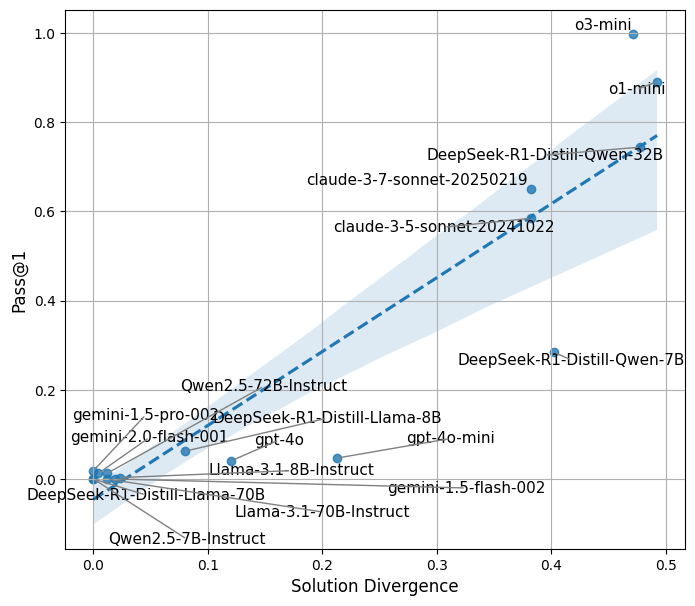}
         \caption{Maze [$\zeta^g_{\pi(h)}$] ($\beta=1.66$)}
         \label{fig:pre_maze_hard}
     \end{subfigure}
     \begin{subfigure}[b]{0.32\textwidth}
         \centering
         \includegraphics[width=\textwidth]{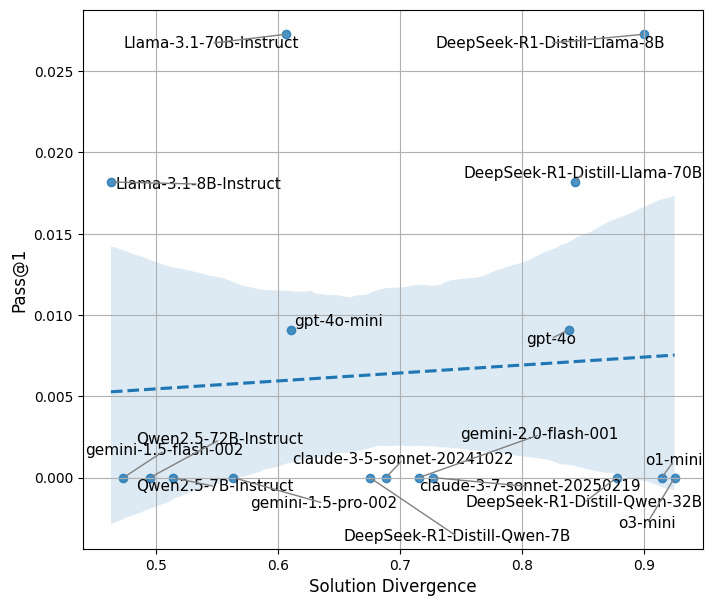}
         \caption{MATH-500 [$\zeta^g_{\pi(h)}$] ($\beta=0.00$)}
         \label{fig:pre_math_hard}
     \end{subfigure}
     \begin{subfigure}[b]{0.32\textwidth}
         \centering
         \includegraphics[width=\textwidth]{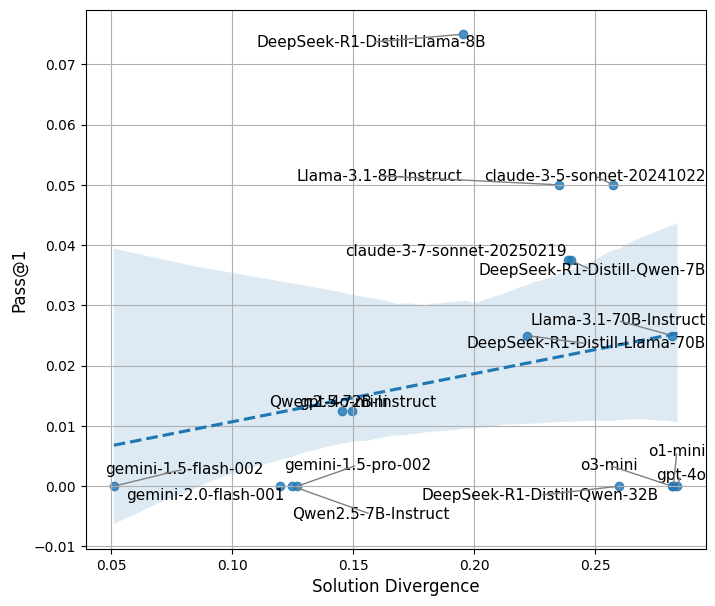}
         \caption{MBPP+ [$\zeta^g_{\pi(h)}$] ($\beta=0.25$)}
         \label{fig:pre_mbpp_hard}
     \end{subfigure}
     \caption{The Relationship of Solution Divergence ($\zeta^{g}_{\pi(e)}$, $\zeta^{g}_{\pi(m)}$, $\zeta^{g}_{\pi(h)}$) to Success Rate (Pass@1) in Maze, Math-500, and MBPP+ Datasets. $\beta$ denotes the slope of the relationship between performance and solution divergence.}
     \label{fig:pre_line_diff}
\end{figure*}

\begin{figure*}[!btph]
     \centering
     \begin{subfigure}[b]{0.32\textwidth}
         \centering
         \includegraphics[width=\textwidth]{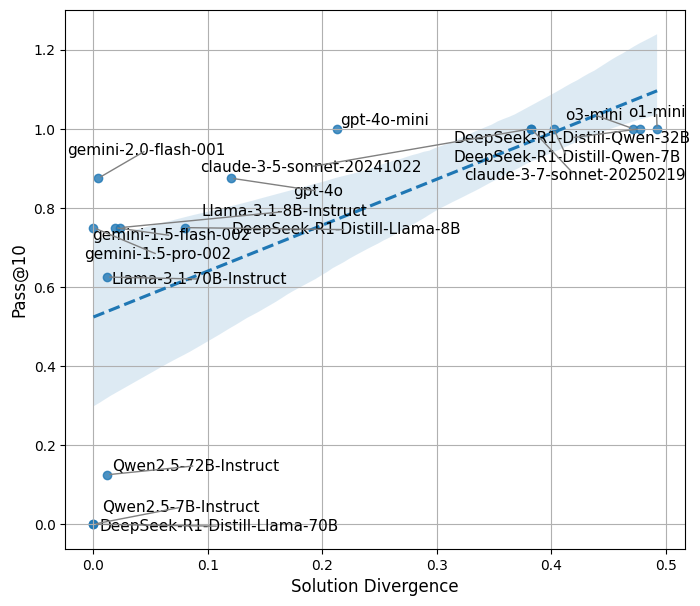}
         \caption{Maze [$\zeta^g_{\pi(e)}$] ($\beta=1.16$)}
         \label{fig:pre_maze_easy_p10}
     \end{subfigure}
     \begin{subfigure}[b]{0.32\textwidth}
         \centering
         \includegraphics[width=\textwidth]{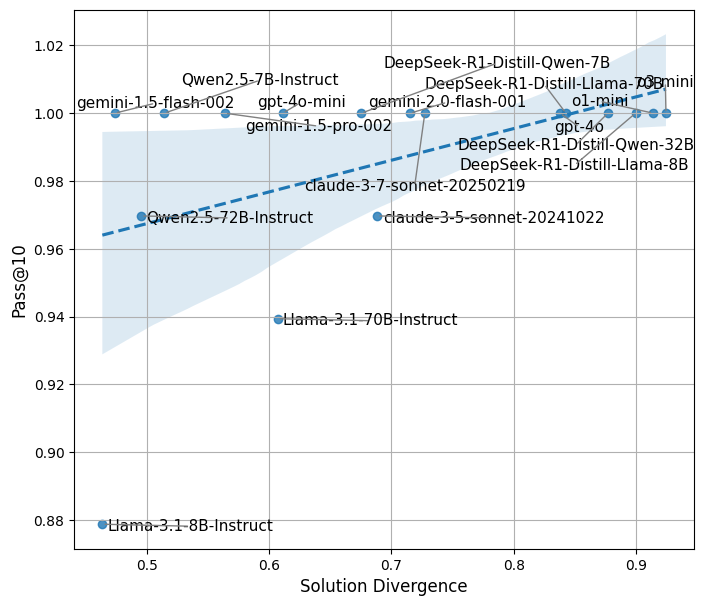}
         \caption{MATH-500 [$\zeta^g_{\pi(e)}$] ($\beta=0.09$)}
         \label{fig:pre_math_easy_p10}
     \end{subfigure}
     \begin{subfigure}[b]{0.32\textwidth}
         \centering
         \includegraphics[width=\textwidth]{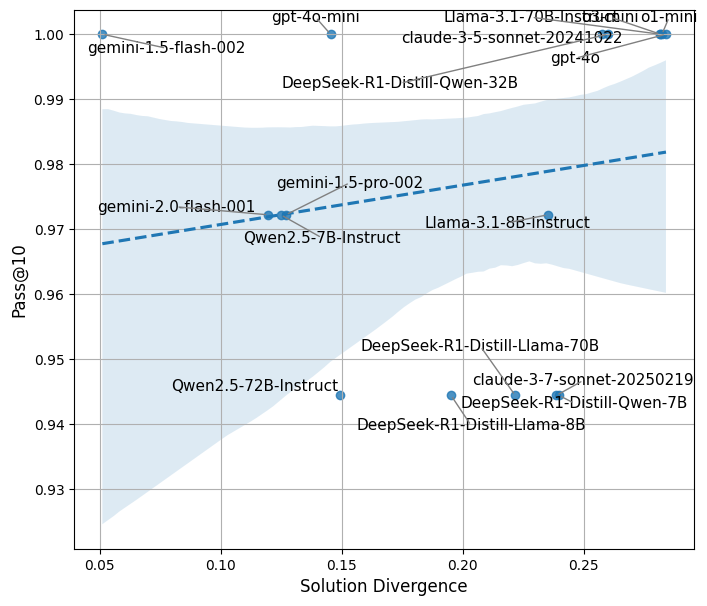}
         \caption{MBPP+ [$\zeta^g_{\pi(e)}$] ($\beta=0.06$)}
         \label{fig:pre_mbpp_easy_p10}
     \end{subfigure}
     \begin{subfigure}[b]{0.32\textwidth}
         \centering
         \includegraphics[width=\textwidth]{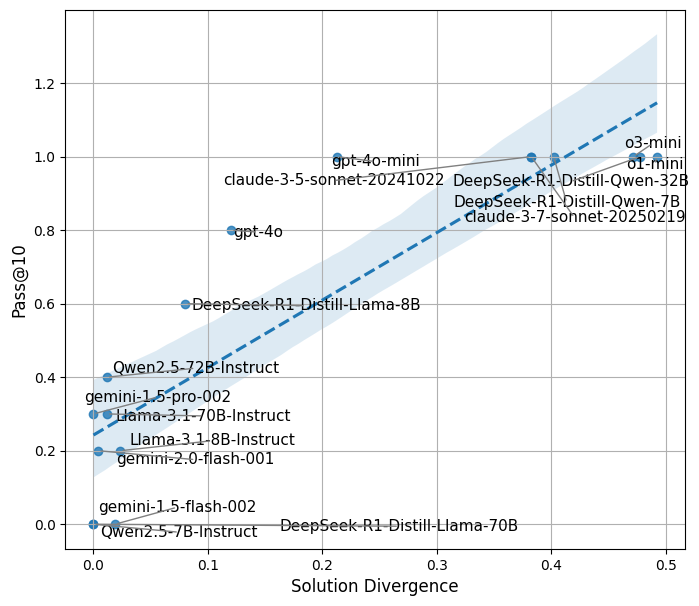}
         \caption{Maze [$\zeta^g_{\pi(m)}$] ($\beta=1.84$)}
         \label{fig:pre_maze_mid_p10}
     \end{subfigure}
     \begin{subfigure}[b]{0.32\textwidth}
         \centering
         \includegraphics[width=\textwidth]{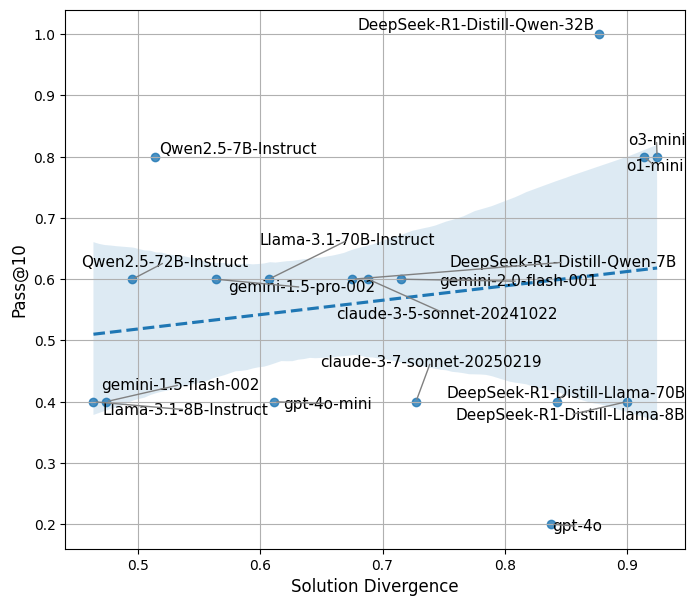}
         \caption{MATH-500 [$\zeta^g_{\pi(m)}$] ($\beta=0.23$)}
         \label{fig:pre_math_mid_p10}
     \end{subfigure}
     \begin{subfigure}[b]{0.32\textwidth}
         \centering
         \includegraphics[width=\textwidth]{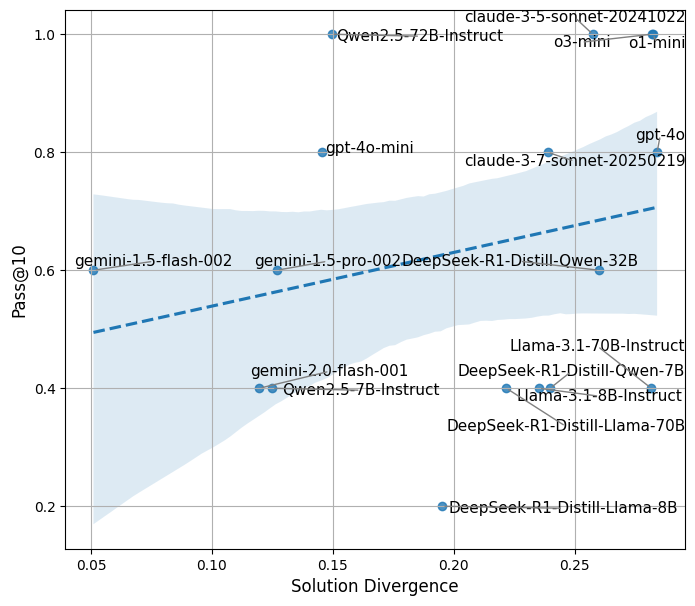}
         \caption{MBPP+ [$\zeta^g_{\pi(m)}$] ($\beta=0.91$)}
         \label{fig:pre_mbpp_mid_p10}
     \end{subfigure}
     \begin{subfigure}[b]{0.32\textwidth}
         \centering
         \includegraphics[width=\textwidth]{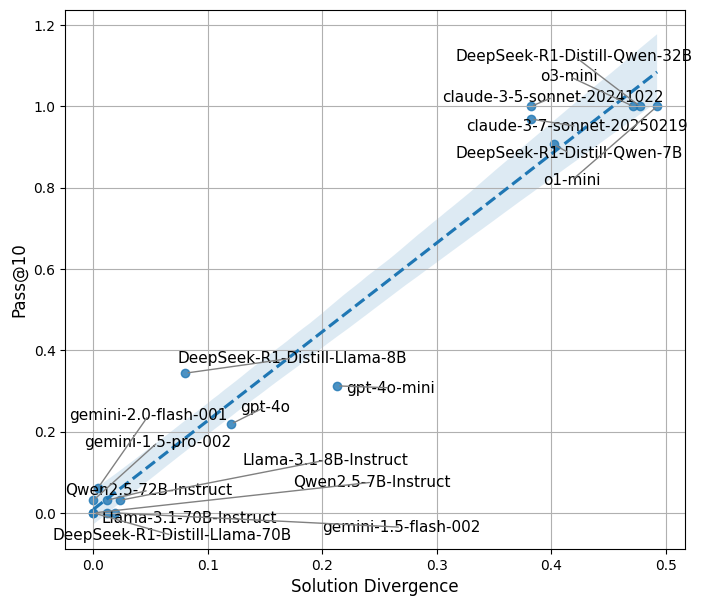}
         \caption{Maze [$\zeta^g_{\pi(h)}$] ($\beta=2.19$)}
         \label{fig:pre_maze_hard_p10}
     \end{subfigure}
     \begin{subfigure}[b]{0.32\textwidth}
         \centering
         \includegraphics[width=\textwidth]{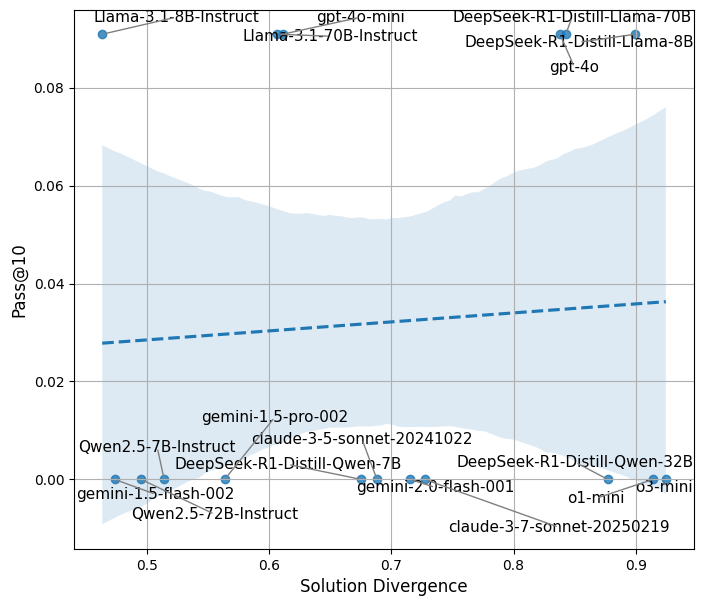}
         \caption{MATH-500 [$\zeta^g_{\pi(h)}$] ($\beta=0.02$)}
         \label{fig:pre_math_hard_p10}
     \end{subfigure}
     \begin{subfigure}[b]{0.32\textwidth}
         \centering
         \includegraphics[width=\textwidth]{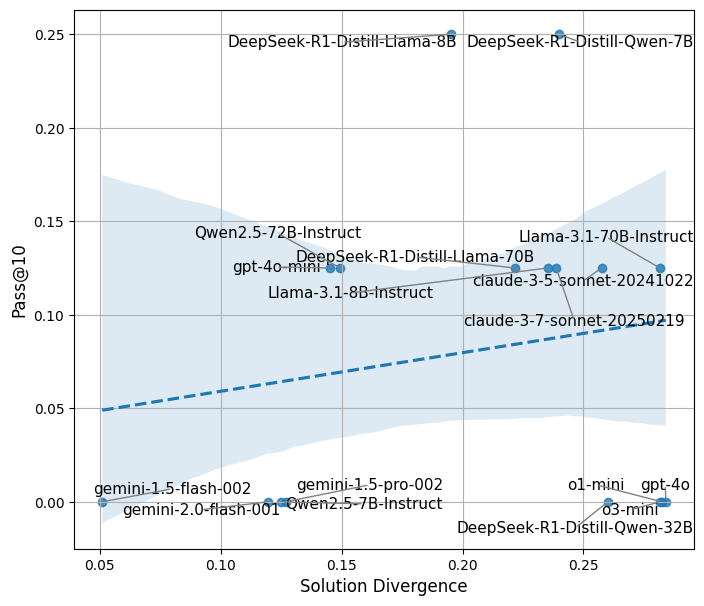}
         \caption{MBPP+ [$\zeta^g_{\pi(h)}$] ($\beta=0.21$)}
         \label{fig:pre_mbpp_hard_p10}
     \end{subfigure}
     \caption{The Relationship of Solution Divergence ($\zeta^{g}_{\pi(e)}$, $\zeta^{g}_{\pi(m)}$, $\zeta^{g}_{\pi(h)}$) to Success Rate (Pass@10) in Maze, Math-500, and MBPP+ Datasets. $\beta$ denotes the slope of the relationship between performance and solution divergence.}
     \label{fig:pre_okay}
\end{figure*}

To examine the alignment between human and LLM problem-solving behavior, we divide each dataset into three difficulty groups, Easy, Medium, and Hard, based on the 33rd and 66th percentiles of the average success rates across all models. For each group, we compute $\zeta_{\pi}^{g}$, denoted as $\zeta_{\pi(e)}^{g}$, $\zeta_{\pi(m)}^{g}$, and $\zeta_{\pi(h)}^{g}$, corresponding to the easy, medium, and hard subsets, respectively, while keeping Pass@1 on the original mixed-difficulty split as the evaluation metric. Fig.~\ref{fig:pre_line_diff} and Fig.~\ref{fig:pre_okay} illustrate the relationship between these group-specific divergence values and Pass@1 and Pass@10, respectively. We observe that the slope $\beta$ of the fitted line is consistently steepest for $\zeta_{\pi(m)}^{g}$, echoing findings from cognitive science~\citep{caviola2018children} that solution divergence is most informative within the medium-difficulty range. This alignment further suggests that divergence serves as a particularly meaningful indicator of capability under moderate problem difficulty. In addition, we observe that only the easy split of MBPP exhibits a negative relationship between solution divergence and performance (Pass@1 and Pass@10). This finding mirrors patterns in human cognition, where, when problems are overly easy, solution divergence becomes less effective at reflecting an individual’s true capability.

\begin{figure}[!btph]
\centering
\begin{tikzpicture}
\draw node[draw=black,fill=black!20,rounded corners,inner sep=2ex,text width=0.45\textwidth,font=\small] {
Given a 2D coordinate system where both the x-axis and y-axis range from 0 to 10 (i.e., units 0, 1, .., 10). 
Consider a point starting at position (0,0). 
The goal is to move this point step by step to the target position: \{target\}. 
During the moving, you cannot pass the following position: \{forbid\}. 
At each step, the point may move only one unit right (r) or one unit up (u). 
Please provide one possible sequence of moves to reach the destination.

The output format should follow this pattern: $\boxed{s\rightarrow r\rightarrow u\rightarrow r\rightarrow \dots\rightarrow e}$, where s indicates the start of the path, and e indicate the end of the path. The steps in between consist only of r and u characters.
If there is no viable path to make the point move to the target position, output as: $\boxed{\times}$;
Do not solve this problem using code or external tools, and avoid including any form of validation or result verification at the end of your response.
};
\end{tikzpicture}
\caption{The example prompt we used to solve the problems of Maze dataset. \{target\} and \{forbid\} are the placeholder for the destination point and block points set, respectively.}
\label{fig:prompt_0}
\end{figure}

\begin{figure}[!btph]
\centering
\begin{tikzpicture}
\draw node[draw=black,fill=black!20,rounded corners,inner sep=2ex,text width=0.45\textwidth,font=\small] {
Please provide a step-by-step solution and final answer to the following question.
Avoid redundant steps, such as restating information from the question or listing pure calculations as independent steps.
Do not include validation or verification at the end of the solution.
\newline
Question:
\{question\}

Format your response as follows:
\newline
Step-by-Step Solution:
Step 1. ...
Step 2. ...
...
Step N. ...
\newline
Final Answer: $\boxed{\text{XXX}}$

(Replace XXX with the final computed value.)
};
\end{tikzpicture}
\caption{The example prompt we used to solve the problems of Math-500 dataset. \{question\} is the placeholder for the question stem text.}
\label{fig:prompt_0}
\end{figure}

\begin{figure}[!btph]
\centering
\begin{tikzpicture}
\draw node[draw=black,fill=black!20,rounded corners,inner sep=2ex,text width=0.45\textwidth,font=\small] {
Complete the Python function below to fulfill the request below.
Please avoid using try, except, or raise statements in your implementation, and focus on achieving the intended functionality.
Request: \{request\}
Return your complete function in the following format:
\begin{verbatim}
```python
{function}
    # start to complete the function here
```
\end{verbatim}
(Replace the comment with your actual implementation.)
};
\end{tikzpicture}
\caption{The example prompt we used to solve the problems of MBPP+. \{request\} is the placeholder for the problem descriptions and \{function\} is the function name required by the execution of the test cases.}
\label{fig:prompt_0}
\end{figure}

\section{Experiment Data Preparation}
\label{appex:exp_data}

In this section, we describe the preparation of training samples for the SFT and RL experiments. Specifically, we use 2,000 and 1,000 questions for math, 98 and 98 for programming, and 250 and 1,000 for logical reasoning in SFT and RL training, respectively. Since our SFT experiments require diverse correct solutions per question, we employ advanced LLMs (e.g., GPT-4o, Gemini-2.5, Claude-3.5) to generate at least 10 distinct correct solutions for each, and the generation prompt for MBPP and Math are shown as Fig.~\ref{fig:mbpp_diverse} and Fig.~\ref{fig:math_diverse}. For Maze, since all viable paths can be enumerated via brute-force search, we directly generate solutions programmatically and randomly sample 10 solutions per question for downstream processing. Then, for each question, we enumerate all 4-solution combinations, compute their solution divergence $\zeta_{q_n}$, and select the subset with the highest divergence as $\mathcal{D}_{\mathcal{S}}^{+}$ and the one with the lowest divergence as $\mathcal{D}_{\mathcal{S}}^{-}$. Aggregating these across all questions yields two version training sets: a high-divergence set $\mathcal{D}_{\text{SFT}}^{+}$ and a low-divergence set $\mathcal{D}_{\text{SFT}}^{-}$. Each version SFT datasets contain 8,000, 392, and 1,000 solutions for math, programming, and logical reasoning, respectively. For validation, we sample 100, 32, and 500 examples from the validation splits of the respective datasets, which are used for both SFT and RL training. For testing, we extend the datasets from the preliminary study: the full 500-question Math-500 set for math, the same 100-question MBPP+ set for programming (as no additional test data are available), and 500 questions for Maze.

\begin{figure}[!btph]
\centering
\begin{tikzpicture}
\draw node[draw=black,fill=black!20,rounded corners,inner sep=2ex,text width=0.45\textwidth,font=\small] {

Provide ten distinct solutions to the single given question.
A reference solution is provided for guidance, but your solutions must be different from the reference.
Each solution must be step-by-step and use a different method from the others.
Avoid redundant steps (e.g., restating the problem or listing bare arithmetic as separate steps).
Do not include validation or verification at the end.

Question
\{question\}

Solution
\{solution\}

Format your response exactly as follows:

Solution 1

Step 1. …
Step 2. …
Step 3. …

Final Answer: $\boxed{\times}$

Solution 2

Step 1. …
…

Final Answer: $\boxed{\times}$

Solution 3

Step 1. …
…

Final Answer: $\boxed{\times}$

Solution 4

Step 1. …
…

Final Answer: $\boxed{\times}$

Now, please start to respond.
};
\end{tikzpicture}
\caption{Example prompt used to generate diverse solutions for SFT training questions of Math dataset.}
\label{fig:math_diverse}
\end{figure}

\begin{figure}[!btph]
\centering
\begin{tikzpicture}
\draw node[draw=black,fill=black!20,rounded corners,inner sep=2ex,text width=0.45\textwidth,font=\small] {
Complete the ten distinct Python functions below to fulfill the request described in the comments.
A reference implementation is provided for guidance, but your solutions must be different from the reference.
Each function should employ a unique approach, and none should rely on try, except, or raise statements. 
Focus on achieving the intended functionality through alternative methods.

\begin{verbatim}
```python
# {request}
# Reference implementation
{solution}
```
\end{verbatim}

Return your complete function in the following format:
\begin{verbatim}
```python
# function 1
{function}

# function 2
{function}

# function 3
{function}

# function 4
{function}
```
\end{verbatim}

Now, please start to respond.
};
\end{tikzpicture}
\caption{Example prompt used to generate diverse solutions for SFT training questions of MBPP+ dataset.}
\label{fig:mbpp_diverse}
\end{figure}

\section{Experiment Setting Details}
\label{appex:exp_setting}

In this section, we introduce the detailed training settings used for our SFT and RL experiment. For both SFT and RL training, we tune the learning rate from $\{2\!\times\!10^{-5},\,1\!\times\!10^{-5},\,8\!\times\!10^{-6},\,5\!\times\!10^{-6}\}$, divergence balancing parameter $\alpha \in \{0,1,2,3,4,5\}$ and set the global batch size to 64. During RL training, we fix the solution set size to $|\mathcal{S}|=8$ per question and set the clipping parameter $\epsilon=0.2$. Both SFT and RL stages are trained for 10 epochs using the Adam optimizer with $\beta_1=0.9$ and $\beta_2=0.999$. During the training, the best performed model on the validation dataset is saved. We implement all the training with the Huggingface TRL packages \footnote{\url{https://huggingface.co/docs/trl/en/index}}.

\section{Experiment with Repeated  Results}
\label{sec:repeat}

As shown in Tab.~\ref{tab:sft} and Tab.~\ref{tab:rl}, the performance gaps between models on metrics such as Pass@1 and Pass@10 narrow in absolute terms. To assess whether these differences remain statistically meaningful, we run inference on the same set of questions 10 times for each trained SFT and RL model using the default decoding settings, and apply a Student’s t-test to evaluate the significance of the average performance differences. The resulting means and standard deviations are reported in Tab.~\ref{tab:reapte_exp}, with statistically significant gaps ($p < 0.05$) marked by an asterisk. From the table, we observe that in most cases the performance differences between models with varying solution divergences remain significant, only 7 out of 24 comparisons are not, demonstrating the consistent influence of solution divergence.

\section{Generation Behavior Statistics}
\label{appx:behavior}

To investigate the influence of solution divergence on model behavior, we report statistics of Qwen2.5-1.5B on generation length and average token entropy during decoding in Tab.~\ref{tab:behavior}. As shown in the table, models trained with $\mathcal{R}_\zeta$ consistently exhibit higher average token entropy than those trained with $\mathcal{R}_s$, reflecting a stronger exploratory tendency induced by solution-divergence-based training. In addition, blindly encouraging higher solution divergence may risk generating redundant or irrelevant content, potentially reducing problem-solving efficiency and increasing generation cost. However,the generation length does not increase substantially after training with $\mathcal{R}_\zeta$. This finding alleviates such concerns and further supports the robustness and effectiveness of our method.

\begin{table}[!btph]
\caption{Generation behavior (length and entropy) of models fine-tuned with rewards $\mathcal{R}_\zeta$ and $\mathcal{R}_s$.}
\label{tab:behavior}
\resizebox{.47\textwidth}{!}{
\begin{tabular}{@{}c|cc|cc@{}}
\toprule
\multirow{2}{*}{\ \ \ \ \ \ \ \ Dataset\ \ \ \ \ \ \ \ } & \multicolumn{2}{c|}{\ \ \ \ \ \ \ \ Length\ \ \ \ \ \ \ \ } & \multicolumn{2}{c}{\ \ \ \ \ \ \ \ Entropy\ \ \ \ \ \ \ \ } \\ \cmidrule(l){2-5} 
 &\ \ \ \  $\mathcal{R}_s$ \ \ \ \ &\ \ \ \  $\mathcal{R}_\zeta$ \ \ \ \ \ \ & \ \ \ \ $\mathcal{R}_s$ \ \ \ \ & \ \ \ \ $\mathcal{R}_\zeta$ \ \ \ \ \\ \midrule
Math-500 & 86 & 98 & 0.176 & 0.701 \\
MBPP+ & 47 & 55 & 0.125 & 0.212 \\ \bottomrule
\end{tabular}}
\end{table}

\begin{table*}[!btph]
\caption{Performance (Pass@1 and Pass@10, in \%) with multiple runs. Results are shown for models with $\mathcal{D}_{\text{SFT}}^-$, $\mathcal{D}_{\text{SFT}}^+$, $\mathcal{R}_s$, and $\mathcal{R}_{\zeta}$. For each metric, we report the mean and std, and statistically significant gaps are marked with *.}
\label{tab:reapte_exp}
\centering
\renewcommand{\arraystretch}{1.2}
\resizebox{\textwidth}{!}{
\begin{tabular}{@{}cc|cccccc|cccccc@{}}
\toprule
\multicolumn{2}{c|}{Model} & \multicolumn{6}{c|}{Llama-3.2-1B} & \multicolumn{6}{c}{Qwen-2.5-1.5B} \\ \midrule
\multicolumn{1}{c|}{Dataset} & Metric & $\mathcal{D}^-_{\text{SFT}}$ & \multicolumn{1}{c|}{$\mathcal{D}^+_{\text{SFT}}$} & \multicolumn{1}{c|}{$\Delta$} & $\mathcal{R}_s$ & \multicolumn{1}{c|}{$\mathcal{R}_{\zeta}$} & $\Delta$ & $\mathcal{D}^-_{\text{SFT}}$ & \multicolumn{1}{c|}{$\mathcal{D}^+_{\text{SFT}}$} & \multicolumn{1}{c|}{$\Delta$} & $\mathcal{R}_s$ & \multicolumn{1}{c|}{$\mathcal{R}_\zeta$} & $\Delta$ \\ \midrule
\multicolumn{1}{c|}{\multirow{2}{*}{Maze}} & Pass@1 & 23.55{\tiny±0.41} & \multicolumn{1}{c|}{23.31{\tiny±0.32}} & \multicolumn{1}{c|}{-0.24} & 25.32{\tiny±0.24} & \multicolumn{1}{c|}{26.57{\tiny±0.37}} & 1.25* & 27.81{\tiny±0.15} & \multicolumn{1}{c|}{26.77{\tiny±0.47}} & \multicolumn{1}{c|}{-1.04*} & 29.73{\tiny±0.25} & \multicolumn{1}{c|}{30.11{\tiny±0.43}} & 0.38\ \ \ \\
\multicolumn{1}{c|}{} & Pass@10 & 34.71{\tiny±0.74} & \multicolumn{1}{c|}{44.71{\tiny±1.09}} & \multicolumn{1}{c|}{10.0*} & 34.88{\tiny±0.83} & \multicolumn{1}{c|}{40.91{\tiny±0.68}} & 6.03* & 39.19{\tiny±0.76} & \multicolumn{1}{c|}{43.45{\tiny±0.84}} & \multicolumn{1}{c|}{4.26*} & 37.15{\tiny±0.55} & \multicolumn{1}{c|}{42.35{\tiny±0.73}} & 5.20* \\ \midrule
\multicolumn{1}{c|}{\multirow{2}{*}{Math-500}} & Pass@1 & 22.78{\tiny±0.16} & \multicolumn{1}{c|}{25.55{\tiny±0.54}} & \multicolumn{1}{c|}{2.77*} & 24.42{\tiny±0.13} & \multicolumn{1}{c|}{25.65{\tiny±0.21}} & 1.23* & 31.65{\tiny±0.39} & \multicolumn{1}{c|}{31.98{\tiny±0.23}} & \multicolumn{1}{c|}{0.33} & 35.62{\tiny±0.15} & \multicolumn{1}{c|}{36.55{\tiny±0.33}} & 0.82* \\
\multicolumn{1}{c|}{} & Pass@10 & 40.25{\tiny±0.87} & \multicolumn{1}{c|}{47.65{\tiny±1.05}} & \multicolumn{1}{c|}{7.40*} & 41.84{\tiny±0.90} & \multicolumn{1}{c|}{41.45{\tiny±1.28}} & -0.39\ \ \ & 51.50{\tiny±1.36} & \multicolumn{1}{c|}{57.28{\tiny±1.43}} & \multicolumn{1}{c|}{5.78*} & 53.90{\tiny±1.66} & \multicolumn{1}{c|}{57.65{\tiny±0.99}} & 3.75* \\ \midrule
\multicolumn{1}{c|}{\multirow{2}{*}{MBPP+}} & Pass@1 & 26.35{\tiny±0.41} & \multicolumn{1}{c|}{28.85{\tiny±1.50}} & \multicolumn{1}{c|}{2.50*} & 33.35{\tiny±0.82} & \multicolumn{1}{c|}{33.45{\tiny±0.47}} & 0.10\ \ & 40.50{\tiny±0.63} & \multicolumn{1}{c|}{40.75{\tiny±1.58}} & \multicolumn{1}{c|}{0.25} & 47.55{\tiny±0.71} & \multicolumn{1}{c|}{48.25{\tiny±0.53}} & 0.70* \\
\multicolumn{1}{c|}{} & Pass@10 & 39.15{\tiny±1.30} & \multicolumn{1}{c|}{50.75{\tiny±1.79}} & \multicolumn{1}{c|}{11.6*} & 44.50{\tiny±0.48} & \multicolumn{1}{c|}{49.52{\tiny±2.28}} & 5.02* & 63.25{\tiny±2.19} & \multicolumn{1}{c|}{63.55{\tiny±1.52}} & \multicolumn{1}{c|}{0.30} & 58.50{\tiny±1.23} & \multicolumn{1}{c|}{62.84{\tiny±0.90}} & 4.34* \\ \bottomrule
\end{tabular}}
\end{table*}

\begin{figure*}[!btph]
     \centering
     \begin{subfigure}[b]{0.32\textwidth}
         \centering
         \includegraphics[width=\textwidth]{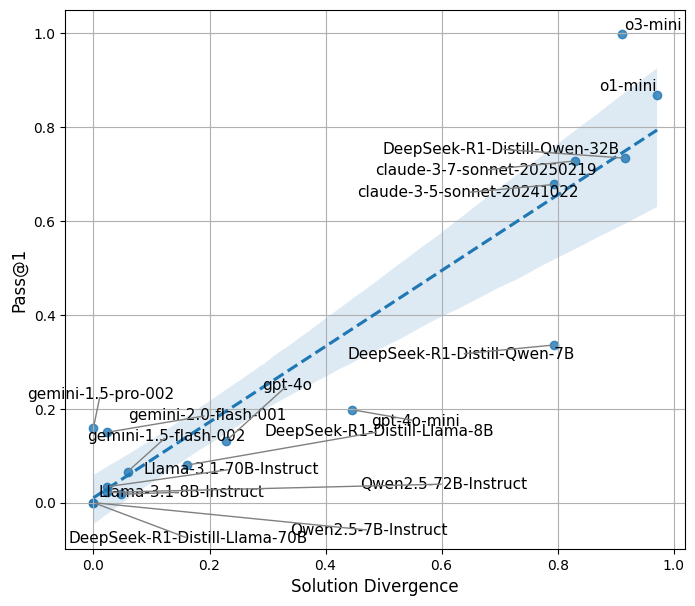}
         \caption{Pass@1 on Maze ($R^2$=.86)}
         \label{fig:pre_maze}
     \end{subfigure}
     \hfill
     \begin{subfigure}[b]{0.32\textwidth}
         \centering
         \includegraphics[width=\textwidth]{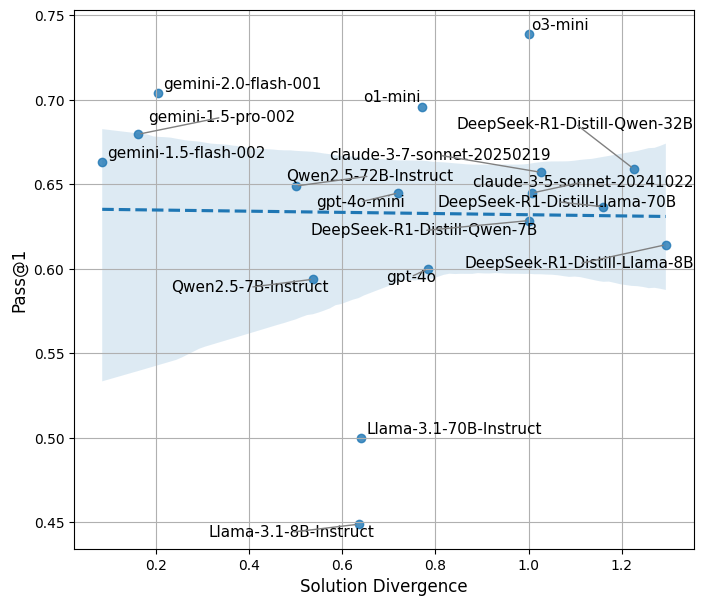}
         \caption{Pass@1 on Math ($R^2$=.01)}
         \label{fig:pre_math}
     \end{subfigure}
     \hfill
     \begin{subfigure}[b]{0.32\textwidth}
         \centering
         \includegraphics[width=\textwidth]{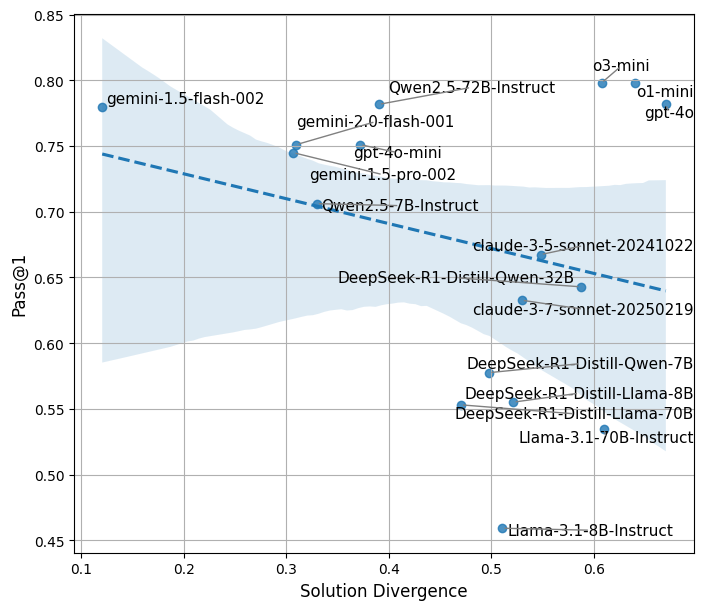}
         \caption{Pass@1 on MBPP ($R^2$=.07)}
         \label{fig:pre_mbpp}
     \end{subfigure}
     \begin{subfigure}[b]{0.32\textwidth}
         \centering
         \includegraphics[width=\textwidth]{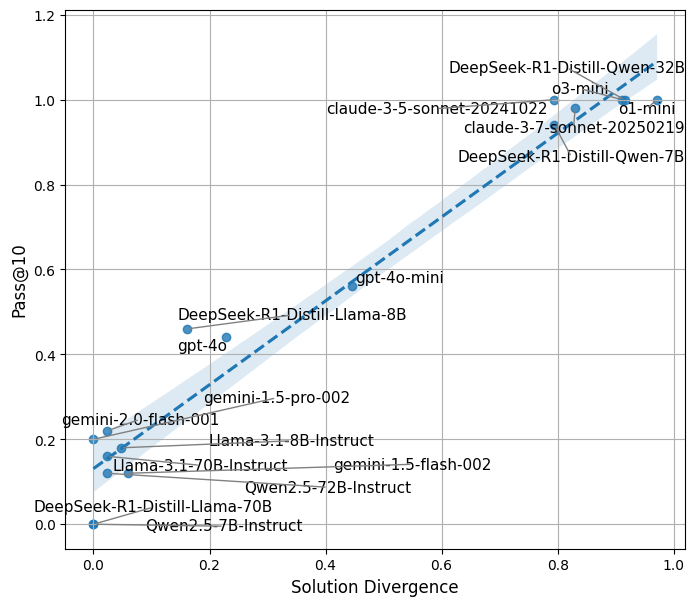}
         \caption{Pass@10 on Maze ($R^2$=.96)}
         \label{fig:pre_maze}
     \end{subfigure}
     \hfill
     \begin{subfigure}[b]{0.32\textwidth}
         \centering
         \includegraphics[width=\textwidth]{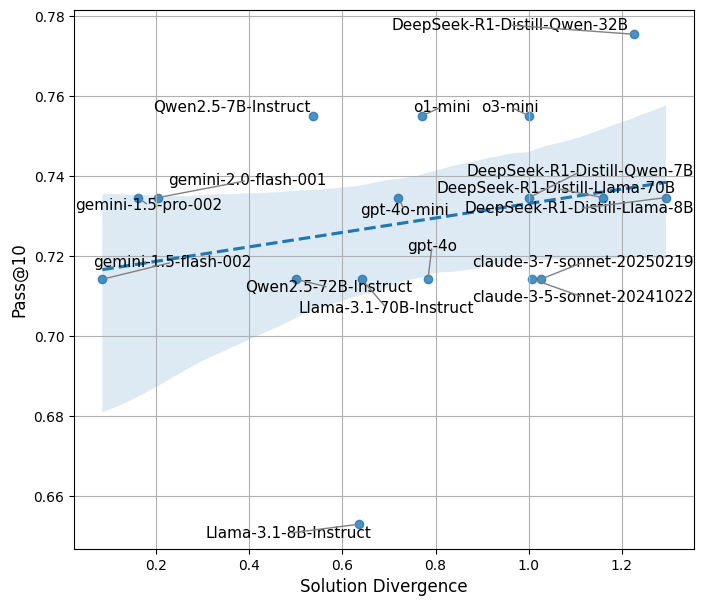}
         \caption{Pass@10 on Math ($R^2$=.06)}
         \label{fig:pre_math}
     \end{subfigure}
     \hfill
     \begin{subfigure}[b]{0.32\textwidth}
         \centering
         \includegraphics[width=\textwidth]{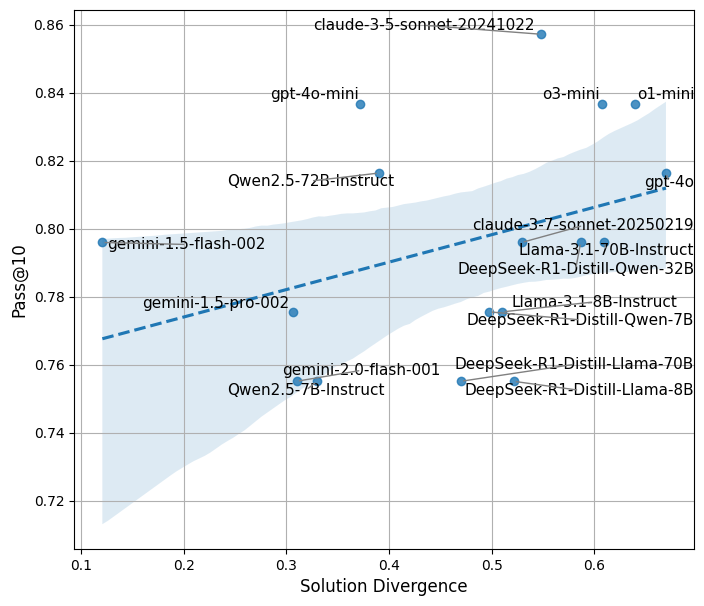}
         \caption{Pass@10 on MBPP ($R^2$=.13)}
         \label{fig:pre_mbpp}
     \end{subfigure}
     \vspace{-2mm}
     \caption{The Relationship of Solution Divergence ($\zeta^{g}_\pi$) to Success Rate (Pass@1 and Pass@10) in Maze, Math-500, and MBPP+ Datasets. For the Maze dataset, $\zeta^{l}_\pi$ is computed using the edit-distance-based metric $d^{(e)}$, while for Math-500 and MBPP+, it is computed using the embedding-based similarity metric $d^{(c)}$.}
     \label{fig:pre_passall}
     \vspace{-5mm}
\end{figure*}

\begin{figure*}[!btph]
     \centering
     \begin{subfigure}[b]{0.32\textwidth}
         \centering
         \includegraphics[width=\textwidth]{figure/Maze-All-Edit-P1.png}
         \caption{Pass@1 on Maze ($R^2$=.86)}
         \label{fig:pre_maze}
     \end{subfigure}
     \hfill
     \begin{subfigure}[b]{0.32\textwidth}
         \centering
         \includegraphics[width=\textwidth]{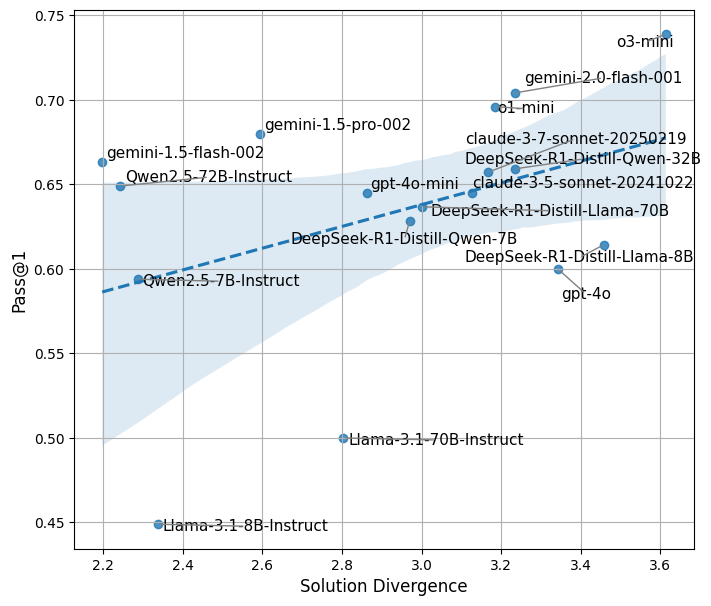}
         \caption{Pass@1 on Math ($R^2$=.16)}
         \label{fig:pre_math}
     \end{subfigure}
     \hfill
     \begin{subfigure}[b]{0.32\textwidth}
         \centering
         \includegraphics[width=\textwidth]{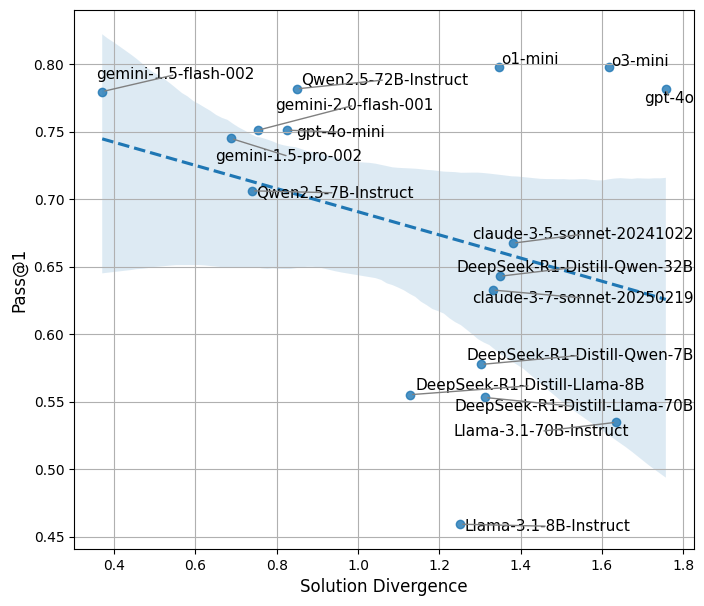}
         \caption{Pass@1 on MBPP ($R^2$=.09)}
         \label{fig:pre_mbpp}
     \end{subfigure}
     \begin{subfigure}[b]{0.32\textwidth}
         \centering
         \includegraphics[width=\textwidth]{figure/Maze-All-Edit-P10.png}
         \caption{Pass@10 on Maze ($R^2$=.96)}
         \label{fig:pre_maze}
     \end{subfigure}
     \hfill
     \begin{subfigure}[b]{0.32\textwidth}
         \centering
         \includegraphics[width=\textwidth]{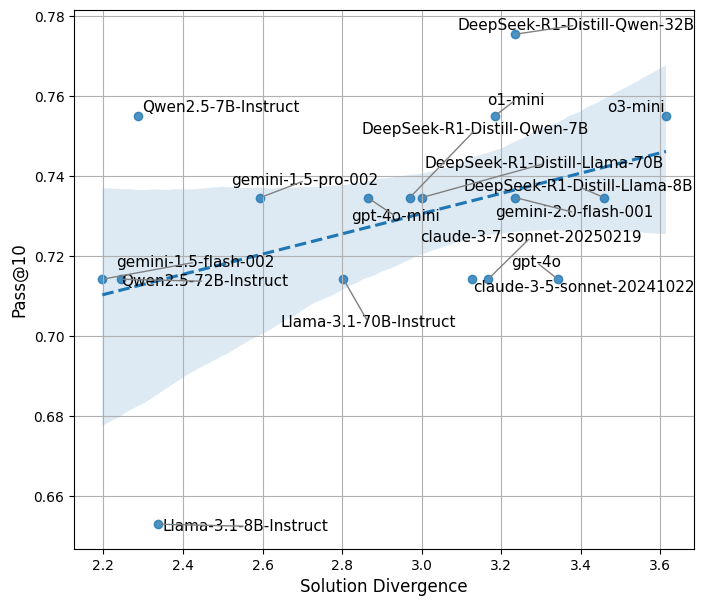}
         \caption{Pass@10 on Math ($R^2$=.16)}
         \label{fig:pre_math}
     \end{subfigure}
     \hfill
     \begin{subfigure}[b]{0.32\textwidth}
         \centering
         \includegraphics[width=\textwidth]{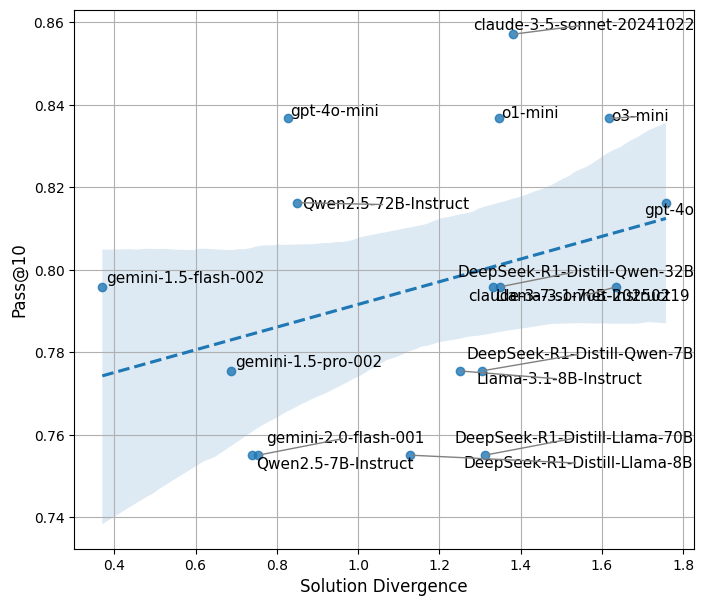}
         \caption{Pass@10 on MBPP ($R^2$=.09)}
         \label{fig:pre_mbpp}
     \end{subfigure}
     \caption{The Relationship of Solution Divergence ($\zeta^{g}_\pi$) to Success Rate (Pass@1 and Pass@10) in Maze, Math-500, and MBPP+ Datasets. For the all dataset, $\zeta^{g}_\pi$ is computed using the edit-distance-based metric $d^{(e)}$.}
     \label{fig:pre_edit}
     \vspace{-5mm}
\end{figure*}

\begin{figure*}[!btph]
     \centering
     \begin{subfigure}[b]{0.32\textwidth}
         \centering
         \includegraphics[width=\textwidth]{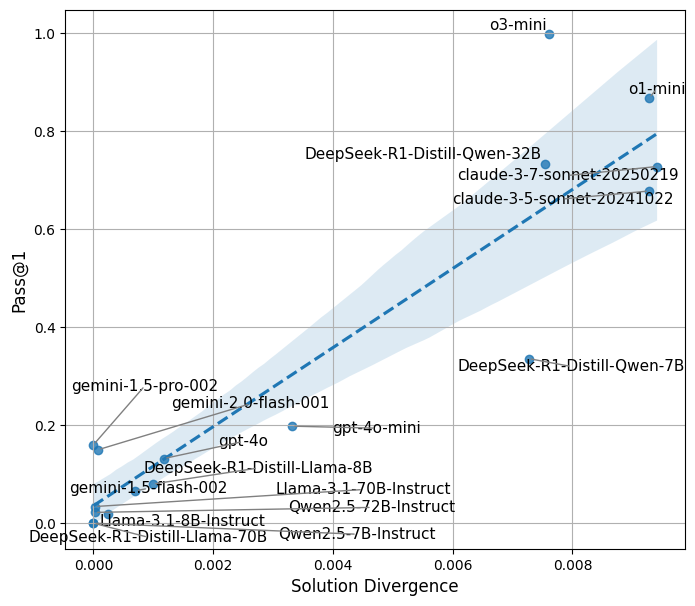}
         \caption{Pass@1 on Maze ($R^2$=.85)}
         \label{fig:pre_maze}
     \end{subfigure}
     \hfill
     \begin{subfigure}[b]{0.32\textwidth}
         \centering
         \includegraphics[width=\textwidth]{figure/Math-All-Embed-P1.png}
         \caption{Pass@1 on Math ($R^2$=.19)}
         \label{fig:pre_math}
     \end{subfigure}
     \hfill
     \begin{subfigure}[b]{0.32\textwidth}
         \centering
         \includegraphics[width=\textwidth]{figure/MBPP-All-Embed-P1.png}
         \caption{Pass@1 on MBPP ($R^2$=.09)}
         \label{fig:pre_mbpp}
     \end{subfigure}
     \begin{subfigure}[b]{0.32\textwidth}
         \centering
         \includegraphics[width=\textwidth]{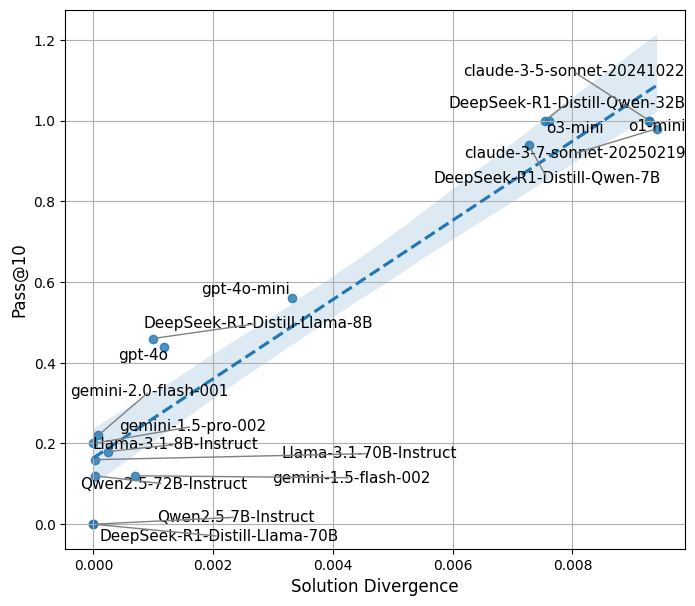}
         \caption{Pass@10 on Maze ($R^2$=.92)}
         \label{fig:pre_maze}
     \end{subfigure}
     \hfill
     \begin{subfigure}[b]{0.32\textwidth}
         \centering
         \includegraphics[width=\textwidth]{figure/Math-All-Embed-P10.png}
         \caption{Pass@10 on Math ($R^2$=.31)}
         \label{fig:pre_math}
     \end{subfigure}
     \hfill
     \begin{subfigure}[b]{0.32\textwidth}
         \centering
         \includegraphics[width=\textwidth]{figure/MBPP-All-Embed-P10.png}
         \caption{Pass@10 on MBPP ($R^2$=.14)}
         \label{fig:pre_mbpp}
     \end{subfigure}
     \caption{The Relationship of Solution Divergence ($\zeta^{g}_\pi$) to Success Rate (Pass@1 and Pass@10) in Maze, Math-500, and MBPP+ Datasets. For the all dataset, $\zeta^{g}_\pi$ is computed using the embedding cosine-distance-based metric $d^{(c)}$.}
     \label{fig:pre_embed}
\end{figure*}